\documentclass{article}

\usepackage{arxiv}
\usepackage{textcomp} 
\usepackage{siunitx}
\usepackage[pdftex, colorlinks=true, linkcolor=snsblue, citecolor=snsgreen, urlcolor=snsred]{hyperref}
\usepackage{url}
\usepackage{booktabs}
\usepackage{amsmath}
\usepackage{amssymb}
\usepackage{amsfonts}
\usepackage{nicefrac}
\usepackage{microtype}
\usepackage{natbib}
\usepackage{siunitx}
\usepackage{mathtools}
\usepackage{algorithm}
\usepackage{algorithmic}
\usepackage[nameinlink]{cleveref}
\usepackage{multirow}
\usepackage{amsopn}
\usepackage{xcolor}

\DeclareMathOperator{\sur}{sur}
\DeclareMathOperator{\gc}{GC}

\colorlet{TestRed}{red!40!white}
\colorlet{TestGreen}{green!40!white}
\definecolor{snsblue}{RGB}{76, 114, 176}
\definecolor{snsgreen}{RGB}{85, 168, 104}
\definecolor{snsred}{RGB}{196, 78, 82}
\definecolor{snspurple}{RGB}{129, 114, 178}
\definecolor{snsyellow}{RGB}{204, 185, 116}
\definecolor{snscyan}{RGB}{100, 181, 205}
\definecolor{snsgray}{gray}{0.75}
\definecolor{smoke}{rgb}{0.96, 0.96, 0.96}
\definecolor{snsorange}{rgb}{0.8666666666666667, 0.5176470588235295, 0.3215686274509804}

\ifpdf
\hypersetup{
  pdftitle={State, global and local parameter estimation using local ensemble Kalman filters: applications to online machine learning of chaotic dynamics},
  pdfauthor={Q. Malartic, A. Farchi and M. Bocquet}
}
\fi

\title{State, global and local parameter estimation using local ensemble Kalman filters: applications to online machine learning of chaotic dynamics}

\author{Quentin Malartic\\
CEREA, École des Ponts and EDF R\&D\\
\^Ile--de--France, France\\
and\\
LMD/IPSL, ENS, PSL Universit\'e, \'Ecole Polytechnique,\\
Institut Polytechnique de Paris, Sorbonne Universit\'e, CNRS,\\
Paris, France\\
\texttt{quentin.malartic@enpc.fr}\\
\AND 
Alban Farchi\\
CEREA, École des Ponts and EDF R\&D\\
\^Ile--de--France, France\\
\texttt{alban.farchi@enpc.fr}\\
\AND
Marc Bocquet\\
CEREA, École des Ponts and EDF R\&D\\
\^Ile--de--France, France\\
\texttt{marc.bocquet@enpc.fr}}

\begin{document}

%\begin{frontmatter}
\maketitle
\begin{abstract}
In a recent methodological paper, we showed how to learn chaotic dynamics along with the state trajectory from sequentially acquired observations, using local ensemble Kalman filters. Here, we more systematically investigate the possibility to use a local ensemble Kalman filter with either covariance localisation or local domains, in order to retrieve the state and a mix of key global and local parameters.
Global parameters are meant to represent the surrogate dynamical core, for instance through a neural network, which is reminiscent of data-driven machine learning of dynamics, while the local parameters typically stand for the forcings of the model. Aiming at joint state and parameter estimation, a family of algorithms for covariance and local domain localisation is proposed.
In particular, we show how to rigorously update global parameters using a local domain ensemble Kalman filter (EnKF) such as the local ensemble transform Kalman filter (LETKF), an inherently local method.
The approach is tested with success on the 40-variable Lorenz model using several of the local EnKF flavors.
A two-dimensional illustration based on a multi-layer Lorenz model is finally provided. It uses radiance-like non-local observations. It features both local domains and covariance localisation in order to learn the chaotic dynamics and the local forcings.
This paper more generally addresses the key question of online estimation of both global and local model parameters.
\end{abstract}
%\end{frontmatter}

%--------------------------------------------------
\section{Introduction}
%--------------------------------------------------
\subsection{Parameter estimation and data-driven techniques for the geosciences}

The recent upheaval generated by machine learning (ML) and in particular deep learning has opened the way to a wealth of data-driven techniques, where not only the state of an observed dynamical system is estimated but also its key dynamical constitutive parameters, if not the full model.
There are by the beginning of 2022, dozens of ML papers in the literature dealing with the problem of estimating the dynamics of a system from observations, even when only focusing on low-order models used in the field of geoscience. The problem can be addressed by typical ML techniques, such as the projection on a regressor frame or basis, random forests, analogs, diffusion maps, reservoir computing, long short-term memory and other neural network approaches \citep[e.g.,][]{brunton2016, lguensat2017, harlim2018, pathak2018a, dueben2018, fablet2018, scher2019, weyn2019, arcomano2020, nadiga2021}. It can also be solved using a conjunction of ML and data assimilation (DA) techniques to exploit noisy and incomplete observations such as those met in realistic geoscience systems \citep{bocquet2019b, brajard2020, bocquet2020, arcucci2021, gottwald2021}.
In the case of high-dimensional systems, the relative lack of information can be compensated by additionally using past trajectories or information on the system such as an approximate model derived from physical laws \citep{wikner2020, brajard2021, farchi2021}.

However, this should not divert us from the even much more abundant contributions focused on the problem of parameter estimation in meteorology, climate science, oceanography, atmospheric chemistry, glaciology, hydrology, solid earth physics, space weather, seismology, etc., using more traditional DA and inverse problem techniques. Compared to the ML view, this part of the geoscience and applied mathematics literature relies much more on a trusted numerical physical model of the system under scrutiny in order to make inferences. 

Nonetheless, the ML scientific tsunami has blurred the frontiers between ML and DA approaches, for the better.
Model error estimation, a classical topic of DA, where the main model is to be corrected through statistical procedures or via parameter estimation techniques, can now be addressed by the addition of an ML based correction with many parameters that need to be learned.
Hence, the coming of ML has pushed the limits of what was traditionally asked of DA, and in particular of DA focused on parameter estimation.

This is the reason why this paper is targeted at filling in some of the current theoretical and algorithmic gaps of DA methods meant for joint state and parameter estimation. The typical applications we have in mind are at that ML/DA frontier where a part or the whole model needs to be learned.
Moreover, following \citet{bocquet2021}, we aim at addressing the difficult objective of learning state and parameters on the fly, \textit{i.e.} online as observations are acquired, using sequential DA techniques such as the ensemble Kalman filter (EnKF) \citep{evensen2009} as an alternative to the variational methods which are more common for parameter and ML problems \citep{farchi2021b}.

\subsection{Local and global model parameters}

In this paper, we assume that the model parameters are not directly observed, which is a common but implicit assumption in the geosciences.
Their inference necessarily indirectly stems from the observation of the state variables.

In an ensemble-based parameter estimation problem, a popular and universal approach consists in augmenting the state vectors to incorporate the parameters \citep{jazwinski1970}. We have adopted it in \citet{bocquet2021} and we will keep doing so here. It was shown that this method also seamlessly blends well with ensemble-variational DA approaches \citep{bocquet2013, bocquet2021}. One can distinguish between two types of parameters, whose nature have a significant impact on the EnKF-based approaches.

First, one can consider \emph{global} parameters, that do not depend on space.
They are typically parameters of the intrinsic physics of the geophysical fluid, of its constituents, or of its dynamics.
However, they are very different from the intrinsically local state variable, leading to substantial theoretical complications, especially when local EnKFs (LEnKFs) based on domain localisation such as the local ensemble transform Kalman filer (LETKF) are used.
This point has been addressed in \citet{bocquet2021} to a large extent, although additional numerical tests and theoretical improvement will be proposed here.

Second, \emph{local} parameters are in a sense simpler to estimate since they are of the same nature as the state variables. However, their number can increase dramatically depending on the number of domains, and yield significantly larger augmented control vectors.
The topic was not addressed in \citet{bocquet2021} but in earlier contributions to the literature as will be discussed in the following section.

In this paper, we will consider both global and local parameters, possibly a mix of them, and develop new EnKF-based algorithms, accounting for the need of localisation in high-dimensional systems. Local parameters could typically represent forcings (radiative forcing, species emissions, etc.), local physical parameters (friction or deposition coefficients) or a Coriolis term while global parameters would represent the parametrised dynamics and micro-physics.

Moreover, in this paper, the parameters are assumed not to depend on time.
This could be induced by an autonomous system, or it could be due to a known and explicit, parametrised dependence on time of, e.g., the forcings, which would be themselves tuned by static parameters. 

\subsection{Parameter estimation techniques in the data assimilation geoscience literature}

Although the literature on parameter estimation based on the EnKF applied to geosciences in high dimensions is vast, the set of available techniques is rather limited. To our knowledge, the state augmentation principle is always used.
\citet{ruiz2013} have written a pedagogical review on parameter estimation with the EnKF, which explains the mechanisms at play. Significant issues with the algorithms arise when local EnKFs are considered. In principle, LEnKFs with covariance localisation (CL) handle global parameters well. However, the extension of the localisation operator to global parameters is not natural, while the addition of local parameters could have an excessive numerical cost.
By contrast, LEnKFs with domain localisation (DL) handle local parameters very well but fail at rigorously estimating global parameters. 

The latter issue, of considerable importance, has been approximately addressed. In \citet{aksoy2006, fertig2009, hu2010},
the global parameters are made local in the DL update step and their local approximations are later averaged in space to form new global parameters (an ad hoc procedure) in order to propagate the ensemble using these updated global parameters.

The former issue where global parameters are estimated with CL LEnKFs, and which requires a definition of the localisation matrix in parameter space as well as the cross-correlations, has been studied by \citet{koyama2010,ruckstuhl2018}.
The authors actually proposed a uniform localisation whenever global parameters are concerned. The localisation matrix associated with the global parameters-state cross-correlation matrix could have its entries set to $1$ (absence of localisation for the global parameters) or to a specific tapering scalar coefficient, which would additionally ensure the positive definiteness of the localisation matrix \citep{ruckstuhl2018}. A generalisation will be proposed in \cref{app:general_zeta}.

In \citet{bocquet2021}, following these first papers, several solutions have been proposed and tested for the CL LEnKF family. Moreover, theoretical solutions were proposed for the DL LEnKF family beyond the approximate solution of \citet{aksoy2006}, but with no numerical tests. These new types of EnKF were termed EnKFs-ML since they were meant to estimate not only the state but also the entire dynamics (through their parameters). \cref{tab:proscons} summarises the adequacy and inadequacy between EnKF families and local/global parameters.

\begin{table}[tb]
    %\rowcolors*{2}{}{}
    \footnotesize
    \centering
    \caption{Adequacy (green) and inadequacy (red) between LEnKF types and the estimation of local, global and mixed parameters. CL refers to covariance localisation and DL refers to domain localisation.}
    \label{tab:proscons}
    \begin{tabular}{lccc}\hline
    LEnKF type &  Global parameters  &  Local parameters & Mixed set of parameters \\\hline
    \multirow{2}{*}{LEnSRF (CL)} & \color{snsgreen}{well suited} & \color{snsgreen}{suited} & \color{snsred}{unclear}\\ & \color{snsred}{localisation in parameter space?} & \color{snsred}{numerically costly} & \color{snsgreen}{solution proposed here} \\\hline
    \multirow{2}{*}{LETKF (DL)} &  \color{snsred}{only approximate} & \color{snsgreen}{well suited} & \color{snsred}{unclear} \\ & \color{snsgreen}{solution proposed here} & & \color{snsgreen}{solution proposed here}  \\\hline
    \end{tabular}
\end{table}

\subsection{Outline}
In \cref{sec:alg}, we will recall, improve and propose parameter estimation techniques, in order to fill the gaps of the geophysical DA parameter estimation literature. In \cref{sec:exp}, the new algorithms will then be evaluated on the Lorenz-96 model with inhomogeneous local forcings. In \cref{sec:2D_experiments},
these algorithms (and combinations thereof) will be tested on a 2D (horizontal and vertical) complex case where radiances are assimilated column-wise, which is reminiscent of a realistic meteorological DA setup. In these experiments, part or the complete model will be learned alongside with the state variables, which represent challenging parameter estimation problems.

\section{Algorithms}
\label{sec:alg}

Following \citet{bocquet2021}, the algorithms derived and tested in this article are based on the augmented EnKF. The main idea is to extend the state vector $\mathbf{x}\in\mathbb{R}^{N_{\mathsf{x}}}$ to $\mathbf{z}\in\mathbb{R}^{N_{\mathsf{z}}}$ containing the state variables and all model parameters. The strength of this approach is that correlations between state variables and parameters will implicitly develop during the forecast steps. Hence the parameters get corrected during the analysis steps even though they are not observed. During the forecast steps, the state variables are updated using the parametrised model, while the parameters follow persistence, \textit{i.e.} are not updated.

Without localisation, the implementation of the ML-counterpart of an EnKF algorithm, the EnKF-ML, is very similar to that of this original algorithm, provided that the observation operator $\boldsymbol{\mathcal{H}}$ has been adjusted for $\mathbf{z}$ instead of $\mathbf{x}$. Hence, in order to avoid divergence, the ensemble size must be strictly larger than the number of neutral and unstable modes of the total (state and parameter) dynamics, equal to the number of neutral and unstable modes of the state dynamics plus the number of influential and independent parameters, as explained by \citet{bocquet2021}. Indeed, the parameter dynamics is entirely neutral since parameters are not updated during the forecast steps \citep{bocquet2021}.

However adding localisation to an EnKF-ML algorithm is not obvious because by definition global parameters cannot be localised. Exploiting the fact that parameters are not observed, we have shown that the EnKF-ML analysis can be written as a two-step process \citep{bocquet2021}: (i) update the state using the observations and (ii) compute the parameter update from the state update using a linear regression based on the ensemble.

More generally, the posterior probability density function (pdf) $p(\mathbf{z}|\mathbf{y})$, which synthesises the analysis problem, can be written
\begin{equation}
    \label{eq:alg-analysis-decomposition}
    p(\mathbf{z}|\mathbf{y}) 
    = p(\mathbf{x}, \mathbf{p}|\mathbf{y}) 
    = p(\mathbf{p}|\mathbf{x},  \mathbf{y})\,p(\mathbf{x}|\mathbf{y}) 
    = p(\mathbf{p}|\mathbf{x})\,p(\mathbf{x}|\mathbf{y}),
\end{equation}
where it has been assumed that the parameter vector $\mathbf{p}$ is independent of the observation vector $\mathbf{y}$ conditional on $\mathbf{x}$. Hence, the analysis can be first carried out on $\mathbf{x}$ by considering the marginal problem on $\mathbf{x}$, integrating out \cref{eq:alg-analysis-decomposition} over $\mathbf{p}$, and later solving the problem on $\mathbf{p}$ once it is solved on $\mathbf{x}$. A consequence of this decomposition is that localisation can be enabled as usual for the state update and disabled for the parameter update. Nevertheless, such an update scheme may be sub-optimal if some of the parameters  are local.

In the following sections, we extend the local EnKF-ML algorithms introduced in \citet{bocquet2021} and apply localisation to the parameter update when possible. The resulting algorithms are called EnKF-HML (for \emph{hybrid} ML) to emphasise the fact that the ML part is partly localised and partly non-localised. Both CL and DL are presented using the example of the ensemble square root Kalman filter (EnSRF) in the first case and of the ensemble transform Kalman filter (ETKF) in the second case \citep{evensen2009}. 

\subsection{Methods and algorithms}

\subsubsection{Partitioning of the augmented state}

We assume that the augmented state $\mathbf{z}\in\mathbb{R}^{N_{\mathsf{z}}}$ is organised as follows:
\begin{equation}
    \label{eq:alg-common-z-organisation}
    \mathbf{z} \triangleq \begin{bmatrix}
    \,\mathbf{x}\, \\ 
    \,\mathbf{p}\, \\ 
    \,\mathbf{q}\,
    \end{bmatrix},
\end{equation}
where $\mathbf{x}\in\mathbb{R}^{N_{\mathsf{x}}}$ is the state of the dynamical system, $\mathbf{p}\in\mathbb{R}^{N_{\mathsf{p}}}$ is the vector of \emph{global} model parameters, and $\mathbf{q}\in\mathbb{R}^{N_{\mathsf{q}}}$ is the vector of \emph{local} model parameters. The ensemble of the filter is a collection of $N_{\mathsf{e}}$ augmented states $\left\{\mathbf{z}_{i}, i=1, \ldots, N_{\mathsf{e}}\right\}$. It is organised column-wise into the augmented ensemble matrix $\mathbf{E}\in\mathbb{R}^{N_{\mathsf{z}}\times N_{\mathsf{e}}}$. The augmented state mean $\bar{\mathbf{z}}\in\mathbb{R}^{N_{\mathsf{z}}}$ and the augmented state perturbation matrix $\mathbf{Z}\in\mathbb{R}^{N_{\mathsf{z}}\times N_{\mathsf{e}}}$ are defined by
\begin{subequations}
    \begin{align}
        \bar{\mathbf{z}} &\triangleq \mathbf{E1}/N_{\mathsf{e}}, \\
        \mathbf{Z} &\triangleq \left(\mathbf{E}-\bar{\mathbf{z}}\mathbf{1}^{\top}\right)/\sqrt{N_{\mathsf{e}}-1},
    \end{align}
\end{subequations}
where $\mathbf{1}\in\mathbb{R}^{N_{\mathsf{e}}}$ is the vector full of ones. 

Following \cref{eq:alg-common-z-organisation}, $\mathbf{E}$, $\bar{\mathbf{z}}$ and $\mathbf{Z}$ can be split according to the state ($\mathsf{x}$), global parameter ($\mathsf{p}$), and local parameter ($\mathsf{q}$) subspaces into
\begin{equation}
    \mathbf{E} = \begin{bmatrix}
    \,\mathbf{E}_{\mathsf{x}}\, \\ 
    \,\mathbf{E}_{\mathsf{p}}\, \\ 
    \,\mathbf{E}_{\mathsf{q}}\,
    \end{bmatrix},
    \quad
    \bar{\mathbf{z}} = \begin{bmatrix}
    \,\bar{\mathbf{x}}\, \\ 
    \,\bar{\mathbf{p}}\, \\ 
    \,\bar{\mathbf{q}}\,
    \end{bmatrix},
    \quad\text{and}\quad
    \mathbf{Z} = \begin{bmatrix}
    \,\mathbf{Z}_{\mathsf{x}}\, \\ 
    \,\mathbf{Z}_{\mathsf{p}}\, \\ 
    \,\mathbf{Z}_{\mathsf{q}}\,
    \end{bmatrix}.
\end{equation}
For these quantities, an "$\mathsf{f}$" superscript is used to refer to the forecast (or prior) value and an "$\mathsf{a}$" superscript is used to refer to the analysis (or posterior) value. Using the same rationale, any matrix $\mathbf{A}\in\mathbb{R}^{N_{\mathsf{z}}\times N_{\mathsf{z}}}$ can also be split into
\begin{equation}
    \label{eq:alg-common-split-A}
    \mathbf{A} = \begin{bmatrix}
    \,\mathbf{A}_{\mathsf{xx}} & \mathbf{A}_{\mathsf{xp}} & \mathbf{A}_{\mathsf{xq}} \, \\ 
    \,\mathbf{A}_{\mathsf{px}} & \mathbf{A}_{\mathsf{pp}} & \mathbf{A}_{\mathsf{pq}}\, \\
    \,\mathbf{A}_{\mathsf{qx}} & \mathbf{A}_{\mathsf{qp}} & \mathbf{A}_{\mathsf{qq}}\,
    \end{bmatrix}.
\end{equation}
In particular, this is the case of the prior error covariance matrix $\mathbf{B}$ and of the localisation matrix $\boldsymbol{\rho}$ for the EnSRF algorithm.

Furthermore, since the model parameters are not observed,
the observation equation can be written \begin{equation}
    \label{eq:alg-observation-equation}
    \mathbf{y} = \boldsymbol{\mathcal{H}}\left(\mathbf{z}\right) = \boldsymbol{\mathcal{H}}_{\mathsf{x}}\left(\mathbf{x}\right),
\end{equation}
where $\boldsymbol{\mathcal{H}}$ is the augmented observation operator and $\boldsymbol{\mathcal{H}}_{\mathsf{x}}$ is the \emph{usual} observation operator (which applies to state only). The tangent linear operators of the maps $\mathbf{x}\mapsto\boldsymbol{\mathcal{H}}_{\mathsf{x}}\left(\mathbf{x}\right)$ and $\mathbf{z}\mapsto\boldsymbol{\mathcal{H}}\left(\mathbf{z}\right)$ are written $\mathbf{H}_{\mathsf{x}}$ and $\mathbf{H}$ and they are related by 
\begin{equation}
    \label{eq:alg-common-def-H}
    \mathbf{H} = \begin{bmatrix}
    \,\mathbf{H}_{\mathsf{x}}\,
    \,\mathbf{0}~\, 
    \,\mathbf{0}\,
    \end{bmatrix}.
\end{equation}

\subsubsection{Matrix square root}

Both the EnSRF and the ETKF are deterministic implementations of the EnKF which rely on a matrix square root. Several definitions of the matrix square root are possible, some of them being non equivalent. In this paper, we use the following definition, chosen, e.g., by \citet{bocquet-2019a, farchi2019}. 

Let $\mathbf{A}$ be a diagonalizable real matrix with non-negative eigenvalues, written $\mathbf{A=GDG}^{-1}$,
where $\mathbf{G}$ is an invertible matrix and $\mathbf{D}$ a diagonal matrix with non-negative entries (the eigenvalues of $\mathbf{A}$). The square root of $\mathbf{A}$, written $\mathbf{A}^{1/2}$, is defined as
\begin{equation}
    \mathbf{A}^{1/2} \triangleq
    \mathbf{GD}^{1/2}\mathbf{G}^{-1},
\end{equation}
where $\mathbf{D}^{1/2}$ is the diagonal matrix containing the square root of the entries of $\mathbf{D}$, \textit{i.e.} the square root of the eigenvalues of $\mathbf{A}$.

\subsection{The ensemble square root Kalman filter}
\label{ssec:alg-ensrf}

\subsubsection{Generic (local) EnSRF analysis}
\label{sssec:alg-generic-ensrf-analysis}

The generic\footnote{The term \emph{generic} is used to describe an algorithm which does not make any distinction between the augmented state variables.} EnSRF analysis is given by the following set of equations:
\begin{subequations}
    \label{eq:alg-generic-ensrf-analysis}
    \begin{align}
        \label{eq:alg-generic-ensrf-mean-update}
        \bar{\mathbf{z}}^{\mathsf{a}} &= \bar{\mathbf{z}}^{\mathsf{f}}+\mathbf{BH}^{\top}\left(\mathbf{R+HBH}^{\top}\right)^{-1}\left(\mathbf{y}-\boldsymbol{\mathcal{H}}\left(\bar{\mathbf{z}}^{\mathsf{f}}\right)\right), \\
        \label{eq:alg-generic-ensrf-pert-update}
        \mathbf{Z}^{\mathsf{a}} &= \left(\mathbf{I+BH}^{\top}\mathbf{R}^{-1}\mathbf{H}\right)^{-1/2}\mathbf{Z}^{\mathsf{f}}.
    \end{align}
\end{subequations}
\Cref{eq:alg-generic-ensrf-mean-update} is known as the \emph{mean update} and \cref{eq:alg-generic-ensrf-pert-update} as the \emph{perturbation update}. In these equations, $\mathbf{B}\in\mathbb{R}^{N_{\mathsf{z}}\times N_{\mathsf{z}}}$ is the prior error covariance matrix, equal in this case to the forecast sample covariance matrix:
\begin{equation}
    \label{eq:alg-generic-ensrf-def-B}
    \mathbf{B} \triangleq \mathbf{Z}^{\mathsf{f}}\left(\mathbf{Z}^{\mathsf{f}}\right)^{\top}.
\end{equation}
Although $\mathbf{I+BH}^{\top}\mathbf{R}^{-1}\mathbf{H}$ may not be symmetric, it is diagonalisable with non-negative eigenvalues \citep[see, for instance,][]{farchi2019}, which makes the matrix square root in \cref{eq:alg-generic-ensrf-pert-update} well defined.

Following \citet{bocquet-2019a}, it can be shown using the matrix shift lemma \citep[see, for instance,][section 6.4.4]{asch2016} that the perturbation update \cref{eq:alg-generic-ensrf-pert-update} is equivalent to
\begin{equation}
    \mathbf{Z}^{\mathsf{a}} = \mathbf{Z}^{\mathsf{f}} - \mathbf{BH}^{\top}\left\{\mathbf{R+HBH}^{\top}+\mathbf{R}\left(\mathbf{I+R}^{-1}\mathbf{HBH}^{\top}\right)^{1/2}\right\}^{-1}\boldsymbol{\mathcal{H}}\left(\mathbf{Z}^{\mathsf{f}}\right),
\end{equation}
where the linear algebra (matrix square root and inverse) is expressed in the observation space, which is usually much smaller than the augmented state space ($N_{\mathsf{y}} \ll N_{\mathsf{z}}$), and where the secant method is used to compute $\boldsymbol{\mathcal{H}}\left(\mathbf{Z}^{\mathsf{f}}\right)$ which stands for
\begin{equation}
    \boldsymbol{\mathcal{H}}\left(\mathbf{Z}^{\mathsf{f}}\right) \triangleq \boldsymbol{\mathcal{H}}\left(\mathbf{E}^{\mathsf{f}}\right)\left(\mathbf{I} - \mathbf{1}\mathbf{1}^\top/N_\mathsf{e}\right)/\sqrt{N_\mathsf{e}-1}.
\end{equation}
We will use this notation throughout the entire manuscript, except in the formal algorithms where its formula is explicit.

Furthermore, the EnSRF analysis \cref{eq:alg-generic-ensrf-analysis} can be written using the following incremental formulation in observation space:
\begin{subequations}
    \label{eq:alg-generic-ensrf-analysis-obs-space}
    \begin{align}
        \Delta\bar{\mathbf{z}} \triangleq \bar{\mathbf{z}}^{\mathsf{a}}-\bar{\mathbf{z}}^{\mathsf{f}} &= \mathbf{BH}^{\top}\left(\mathbf{R+HBH}^{\top}\right)^{-1}\left(\mathbf{y}-\boldsymbol{\mathcal{H}}\left(\bar{\mathbf{z}}^{\mathsf{f}}\right)\right), \\
        \Delta\mathbf{Z} \triangleq
        \mathbf{Z}^{\mathsf{a}}-\mathbf{Z}^{\mathsf{f}} &= -\mathbf{BH}^{\top}\left\{\mathbf{R+HBH}^{\top}+\mathbf{R}\left(\mathbf{I+R}^{-1}\mathbf{HBH}^{\top}\right)^{1/2}\right\}^{-1}\boldsymbol{\mathcal{H}}\left(\mathbf{Z}^{\mathsf{f}}\right).
    \end{align}
\end{subequations}

This update can be further simplified if $\mathbf{R}^{-1/2}$ is easy to compute, for instance if $\mathbf{R}$ is diagonal as is often assumed in the geosciences. In this case, let us introduce the ancillary matrix $\mathbf{T}_{\mathsf{y}}\in\mathbb{R}^{N_{\mathsf{y}}\times N_{\mathsf{y}}}$ defined as
\begin{equation}
    \label{eq:alg-ensrf-ml-def-T}
    \mathbf{T}_{\mathsf{y}} \triangleq \mathbf{I+R}^{-1/2}\mathbf{HBH}^{\top}\mathbf{R}^{-1/2}.
\end{equation}
Introducing $\mathbf{T}_{\mathsf{y}}$ in \cref{eq:alg-generic-ensrf-analysis-obs-space} yields
\begin{subequations}
    \begin{align}
        \label{eq:alg-generic-ensrf-mean-update-incr}
        \Delta\bar{\mathbf{z}} &=
        \mathbf{BH}^{\top}\mathbf{R}^{-1/2}\mathbf{T}_{\mathsf{y}}^{-1}\mathbf{R}^{-1/2}\left(\mathbf{y}-\boldsymbol{\mathcal{H}}\left(\bar{\mathbf{z}}^{\mathsf{f}}\right)\right),\\
        \label{eq:alg-generic-ensrf-pert-update-incr}
        \Delta\mathbf{Z} &= -\mathbf{BH}^{\top}\mathbf{R}^{-1/2}\left(\mathbf{T}_{\mathsf{y}}+\mathbf{T}_{\mathsf{y}}^{1/2}\right)^{-1}\mathbf{R}^{-1/2}\boldsymbol{\mathcal{H}}\left(\mathbf{Z}^{\mathsf{f}}\right),
    \end{align}
\end{subequations}
where the linear algebra operators now apply to symmetric matrices only.

Finally, CL can be included in the analysis by replacing \cref{eq:alg-generic-ensrf-def-B} with
\begin{equation}
    \mathbf{B} = \boldsymbol{\rho}\circ\left[\mathbf{Z}^{\mathsf{f}}\left(\mathbf{Z}^{\mathsf{f}}\right)^{\top}\right],
\end{equation}
where $\boldsymbol{\rho}\in\mathbb{R}^{N_{\mathsf{z}}\times N_{\mathsf{z}}}$ is the localisation matrix, a correlation matrix which depends on the geometry\footnote{By geometry, we mean here the number of variables and their position in space.} of all variables, and $\circ$ is the Schur/Hadamard product. The resulting analysis is called the local EnSRF (LEnSRF).

In the following sections, we show how the EnSRF analysis (both global and local) can be efficiently implemented when the augmented state contains model parameters. In \cref{sssec:alg-ensrf-ml-analysis}, we only consider global model parameters, repeating \citet{bocquet2021}, and in \cref{sssec:alg-ensrf-hml-analysis}, we consider the general case with both global and local model parameters.

\subsubsection{The (local) EnSRF-ML analysis}
\label{sssec:alg-ensrf-ml-analysis}

Let us start with global parameters only (\textit{i.e.} $N_{\mathsf{q}}=0$). Following \citet{bocquet2021}, it is possible to separate state and parameter update in the analysis to make it more efficient. 

To do this, we split $\mathbf{B}$ according to the state and parameter subspaces as in \cref{eq:alg-common-split-A}. The ancillary matrix $\mathbf{T}_{\mathsf{y}}$ is equal to
\begin{equation}
    \label{eq:alg-ensrf-ml-Ty}
    \mathbf{T}_{\mathsf{y}} = \mathbf{I+R}^{-1/2}\mathbf{H}_{\mathsf{x}}\mathbf{B}_{\mathsf{xx}}\mathbf{H}^{\top}_{\mathsf{x}}\mathbf{R}^{-1/2}.
\end{equation}
Let us now introduce the additional ancillary variables $\mathbf{u}_{\mathsf{x}}\in\mathbb{R}^{N_{\mathsf{x}}}$ and $\mathbf{U}_{\mathsf{x}}\in\mathbb{R}^{N_{\mathsf{x}}\times N_{\mathsf{e}}}$ defined as
\begin{subequations}
    \label{eq:alg-ensrf-ml-def-u}
    \begin{align}
        \label{eq:alg-ensrf-ml-def-u-mean}
        \mathbf{u}_{\mathsf{x}} &\triangleq
        \mathbf{H}^{\top}_{\mathsf{x}}\mathbf{R}^{-1/2}\mathbf{T}_{\mathsf{y}}^{-1}\mathbf{R}^{-1/2}\left(\mathbf{y}-\boldsymbol{\mathcal{H}}\left(\bar{\mathbf{z}}^{\mathsf{f}}\right)\right),\\
        \label{eq:alg-ensrf-ml-def-u-pert}
        \mathbf{U}_{\mathsf{x}} &\triangleq -\mathbf{H}^{\top}_{\mathsf{x}}\mathbf{R}^{-1/2}\left(\mathbf{T}_{\mathsf{y}}+\mathbf{T}_{\mathsf{y}}^{1/2}\right)^{-1}\mathbf{R}^{-1/2}\boldsymbol{\mathcal{H}}\left(\mathbf{Z}^{\mathsf{f}}\right).
    \end{align}
\end{subequations}
With these definitions, the mean update \cref{eq:alg-generic-ensrf-mean-update-incr} becomes
\begin{subequations}
    \begin{align}
        \label{eq:alg-ensrf-ml-state-mean-update}
        \Delta\bar{\mathbf{x}} &= \mathbf{B}_{\mathsf{xx}}\mathbf{u}_{\mathsf{x}}, \\
        \label{eq:alg-ensrf-ml-param-mean-update}
        \Delta\bar{\mathbf{p}} &= \mathbf{B}_{\mathsf{px}}\mathbf{u}_{\mathsf{x}},
    \end{align}
\end{subequations}
and the perturbation update \cref{eq:alg-generic-ensrf-pert-update-incr} becomes
\begin{subequations}
    \begin{align}
        \label{eq:alg-ensrf-ml-state-pert-update}
        \Delta\mathbf{Z}_{\mathsf{x}} &= \mathbf{B}_{\mathsf{xx}}\mathbf{U}_{\mathsf{x}}, \\
        \label{eq:alg-ensrf-ml-param-pert-update}
        \Delta\mathbf{Z}_{\mathsf{p}} &= \mathbf{B}_{\mathsf{px}}\mathbf{U}_{\mathsf{x}}.
    \end{align}
\end{subequations}

Assuming that $\mathbf{B}_{\mathsf{xx}}$ is invertible, the parameter update formulae \cref{eq:alg-ensrf-ml-param-mean-update,eq:alg-ensrf-ml-param-pert-update}, can be written
\begin{subequations}
    \label{eq:alg-ensrf-ml-param-update-original}
    \begin{align}
        \Delta\bar{\mathbf{p}} &= 
        \mathbf{B}_{\mathsf{px}}\mathbf{B}^{-1}_{\mathsf{xx}}\Delta\bar{\mathbf{x}}, \\
        \Delta\mathbf{Z}_{\mathsf{p}} &= 
        \mathbf{B}_{\mathsf{px}}\mathbf{B}^{-1}_{\mathsf{xx}}\Delta\mathbf{Z}_{\mathsf{x}},
    \end{align}
\end{subequations}
which is the original parameter update derived by \citet{bocquet2021}. Note that $\mathbf{B}^{-1}_{\mathsf{xx}}$ does not need to be computed since it applies to $N_{\mathsf{e}} \ll N_{\mathsf{x}}$ vectors and only requires the solution of a linear system of equations with $N_{\mathsf{e}}N_{\mathsf{x}}$ unknowns. 

At this point, it is important to realise that the state-wise update \cref{eq:alg-ensrf-ml-state-mean-update,eq:alg-ensrf-ml-state-pert-update} is the usual EnSRF analysis while the parameter update \cref{eq:alg-ensrf-ml-param-update-original} is a regression of the state update into the parameter subspace. Such regression is very general and can be used regardless of the state update method. In our specific case however, the introduction of the ancillary variables $\mathbf{u}_{\mathsf{x}}$ and $\mathbf{U}_{\mathsf{x}}$ permits us to bypass the matrix multiplication by $\mathbf{B}^{-1}_{\mathsf{xx}}$. In a way, one can think of $\mathbf{u}_{\mathsf{x}}$ and $\mathbf{U}_{\mathsf{x}}$ as the uncorrelated increments. Moreover, note that the entire analysis does not depend on $\mathbf{B}_{\mathsf{pp}}$, neither explicitly nor implicitly.

When using CL, as for $\mathbf{B}$, we split $\boldsymbol{\rho}$ in such a way that
\begin{subequations}
    \label{eq:alg-ensrf-ml-local-B}
    \begin{align}
        \mathbf{B}_{\mathsf{xx}} &= \boldsymbol{\rho}_{\mathsf{xx}}\circ\left[\mathbf{Z}^{\mathsf{f}}_{\mathsf{x}}\left(\mathbf{Z}^{\mathsf{f}}_{\mathsf{x}}\right)^{\top}\right], \\
        \mathbf{B}_{\mathsf{px}} &=  \boldsymbol{\rho}_{\mathsf{px}}\circ\left[\mathbf{Z}^{\mathsf{f}}_{\mathsf{p}}\left(\mathbf{Z}^{\mathsf{f}}_{\mathsf{x}}\right)^{\top}\right] = \mathbf{B}^{\top}_{\mathsf{xp}}, \\
        \mathbf{B}_{\mathsf{pp}} &= \boldsymbol{\rho}_{\mathsf{pp}}\circ\left[\mathbf{Z}^{\mathsf{f}}_{\mathsf{p}}\left(\mathbf{Z}^{\mathsf{f}}_{\mathsf{p}}\right)^{\top}\right].
    \end{align}
\end{subequations}
The localisation matrix for the state subspace $\boldsymbol{\rho}_{\mathsf{xx}}$ is the usual localisation matrix. It almost certainly makes $\mathbf{B}_{\mathsf{xx}}$ positive definite, and in particular invertible. 
The localisation matrix for the state-parameter cross subspace $\boldsymbol{\rho}_{\mathsf{px}}$ has to be row-wise uniform because the parameters are global.
Hence it is of the form 
\begin{equation}
\label{eq:general-cross-localisation}
    \boldsymbol{\rho}_{\mathsf{px}}=\boldsymbol{\zeta}_{\mathsf{p}}\mathbf{1}^\top_{\mathsf{x}},
\end{equation}
where $\mathbf{1}_{\mathsf{x}}\in\mathbb{R}^{N_{\mathsf{x}}}$ is the vector full of ones and where $\boldsymbol{\zeta_{\mathsf{p}}
} \in \mathbb{R}^{N_{\mathsf{p}}}$ is a vector of algorithmic parameters, one for each model parameter, which is more general than what was suggested in \citet{bocquet2021}.
Nonetheless, for the sake of simplicity, we choose in the following this vector to be uniform as in \citet{bocquet2021}, such that $\boldsymbol{\rho}_{\mathsf{px}}= \zeta_{\mathsf{p}}\boldsymbol{\Pi}_{\mathsf{px}}$, where $\zeta_{\mathsf{p}}$ is a scalar algorithmic parameter (please see \cref{app:general_zeta} for the multivariate generalisation). Looking at \cref{eq:alg-ensrf-ml-param-mean-update,eq:alg-ensrf-ml-param-pert-update}, we see that $\zeta_{\mathsf{p}}$ tapers the parameter update in a linear way: using $\zeta_{\mathsf{p}}=1$ does not alter the parameter update while using $\zeta_{\mathsf{p}}=0$ entirely disables the parameter update. For this reason, $\zeta_{\mathsf{p}}$ is called the \emph{tapering} parameter. For simplicity and to emphasise the role of the tapering, we assume that $\boldsymbol{\rho}_{\mathsf{px}}=\boldsymbol{\Pi}_{\mathsf{px}}$ and we introduce $\zeta_{\mathsf{p}}$ directly into the parameter update, which is now written 
\begin{subequations}
    \label{eq:alg-ensrf-ml-param-update-tap}
    \begin{align}
        \Delta\bar{\mathbf{p}} &= \zeta_{\mathsf{p}}\mathbf{B}_{\mathsf{px}}\mathbf{u}_{\mathsf{x}}, \\
        \Delta\mathbf{Z}_{\mathsf{p}} &= \zeta_{\mathsf{p}}\mathbf{B}_{\mathsf{px}}\mathbf{U}_{\mathsf{x}}.
    \end{align}
\end{subequations}
Finally, since $\mathbf{B}_{\mathsf{pp}}$ is not used during the analysis, $\boldsymbol{\rho}_{\mathsf{pp}}$ does not need to be specified. This means that potential spurious correlations between global parameters are not mitigated, but this is not problematic because such correlations have no effect on the analysis ensemble. Nevertheless, since the localisation matrix $\boldsymbol{\rho}$ is by assumption a correlation matrix, it must be symmetric and positive definite. This means that the tapering parameter $\zeta_{\mathsf{p}}$ cannot take arbitrary values \citep{ruckstuhl2018}. See \citet{bocquet2021} for a detailed interpretation of $\zeta_{\mathsf{p}}$.

For completeness, let us mention that the update \cref{eq:alg-ensrf-ml-param-update-original} is also valid without localisation when $\mathbf{B}_{\mathsf{xx}}$ is not invertible (\textit{i.e.} when $N_{\mathsf{e}}\leq N_{\mathsf{x}}+1$) provided that we replace the inverse by the Moore--Penrose pseudo-inverse. Indeed in this case, as proven in \citet{bocquet2021}, the parameter update can be written
\begin{subequations}
    \label{eq:alg-ensrf-ml-param-update-original-pinv}
    \begin{align}
        \Delta\bar{\mathbf{p}} &= \mathbf{Z}^{\mathsf{f}}_{\mathsf{p}}\left(\mathbf{Z}^{\mathsf{f}}_{\mathsf{x}}\right)^{+}\Delta\bar{\mathbf{x}}, \\
        \Delta\mathbf{Z}_{\mathsf{p}} &= \mathbf{Z}^{\mathsf{f}}_{\mathsf{p}}\left(\mathbf{Z}^{\mathsf{f}}_{\mathsf{x}}\right)^{+}\Delta\mathbf{Z}_{\mathsf{x}},
    \end{align} 
\end{subequations}
where the Moore--Penrose pseudo-inverse is indicated by a $+$ superscript. Realising that 
\begin{equation}
    \label{eq:alg-ensrf-ml-pinv-equivalence}
    \mathbf{Z}^{\mathsf{f}}_{\mathsf{p}}\left(\mathbf{Z}^{\mathsf{f}}_{\mathsf{x}}\right)^{+} = \mathbf{Z}^{\mathsf{f}}_{\mathsf{p}}\left(\mathbf{Z}^{\mathsf{f}}_{\mathsf{x}}\right)^{\top}\left[\mathbf{Z}^{\mathsf{f}}_{\mathsf{x}}\left(\mathbf{Z}^{\mathsf{f}}_{\mathsf{x}}\right)^{\top}\right]^{+} = \mathbf{B}_{\mathsf{px}}\mathbf{B}^{+}_{\mathsf{xx}},
\end{equation}
we conclude that the update \cref{eq:alg-ensrf-ml-param-update-original-pinv} is equivalent to \cref{eq:alg-ensrf-ml-param-update-original} upon replacing the inverse by the Moore--Penrose pseudo-inverse.

\subsubsection{The (local) EnSRF-HML analysis}
\label{sssec:alg-ensrf-hml-analysis}

We now extend the EnSRF-ML analysis to the case where both global and local parameters are estimated. For this problem, we keep the state update and the global parameter update of the EnSRF-ML analysis, namely \cref{eq:alg-ensrf-ml-state-mean-update,eq:alg-ensrf-ml-state-pert-update,eq:alg-ensrf-ml-param-update-tap}, and an update for the local parameters needs to be provided. 

Following the arguments of \cref{sssec:alg-ensrf-ml-analysis}, we choose to write the local parameter update as
\begin{subequations}
    \label{eq:alg-ensrf-hml-local-param-update-original}
    \begin{align}
        \Delta\bar{\mathbf{q}} &= \mathbf{B}_{\mathsf{qx}}\mathbf{u}_{\mathsf{x}}, \\
        \Delta\mathbf{Z}_{\mathsf{q}} &= \mathbf{B}_{\mathsf{qx}}\mathbf{U}_{\mathsf{x}}.
    \end{align}
\end{subequations}
Without localisation, there is no distinction between local and global parameters. When using CL, in addition to \cref{eq:alg-ensrf-ml-local-B}, we have
\begin{subequations}
    \begin{align}
        \mathbf{B}_{\mathsf{qx}} &=  \boldsymbol{\rho}_{\mathsf{qx}}\circ\left[\mathbf{Z}^{\mathsf{f}}_{\mathsf{q}}\left(\mathbf{Z}^{\mathsf{f}}_{\mathsf{x}}\right)^{\top}\right] = \mathbf{B}^{\top}_{\mathsf{xq}}, \\
        \mathbf{B}_{\mathsf{qp}} &=  \boldsymbol{\rho}_{\mathsf{qp}}\circ\left[\mathbf{Z}^{\mathsf{f}}_{\mathsf{q}}\left(\mathbf{Z}^{\mathsf{f}}_{\mathsf{p}}\right)^{\top}\right] = \mathbf{B}^{\top}_{\mathsf{pq}}, \\
        \mathbf{B}_{\mathsf{qq}} &= \boldsymbol{\rho}_{\mathsf{qq}}\circ\left[\mathbf{Z}^{\mathsf{f}}_{\mathsf{q}}\left(\mathbf{Z}^{\mathsf{f}}_{\mathsf{q}}\right)^{\top}\right].
    \end{align}
\end{subequations}
Again, since $\mathbf{B}_{\mathsf{qp}}$ and $\mathbf{B}_{\mathsf{qq}}$ are not used during the analysis, $\boldsymbol{\rho}_{\mathsf{qp}}$ and $\boldsymbol{\rho}_{\mathsf{qq}}$ do not need to be specified. The localisation matrix for the state-local parameter cross subspace $\boldsymbol{\rho}_{\mathsf{qx}}$ has to reflect the geometry of the local parameters and state variables, contrary to $\boldsymbol{\rho}_{\mathsf{px}}$ which, as explained in \cref{sssec:alg-ensrf-ml-analysis}, is bound to be row-wise uniform. This point is important since specifying $\boldsymbol{\rho}_{\mathsf{qx}}$ is the only way to include localisation in the local parameter update.

Finally, it is possible to normalise $\boldsymbol{\rho}_{\mathsf{qx}}$ into $\widehat{\boldsymbol{\rho}}_{\mathsf{qx}}$ using its largest value: $\boldsymbol{\rho}_{\mathsf{qx}}=\zeta_{\mathsf{q}}\widehat{\boldsymbol{\rho}}_{\mathsf{qx}}$. It turns out that $\zeta_{\mathsf{q}}$, defined as the largest value of $\boldsymbol{\rho}_{\mathsf{qx}}$, has the same role in the local parameter update \cref{eq:alg-ensrf-hml-local-param-update-original} than the tapering parameter $\zeta_{\mathsf{p}}$ in the global parameter update \cref{eq:alg-ensrf-ml-param-mean-update,eq:alg-ensrf-ml-param-pert-update}. Therefore, as in the previous section, we assume that $\boldsymbol{\rho}_{\mathsf{qx}}=\widehat{\boldsymbol{\rho}}_{\mathsf{qx}}$ and we introduce $\zeta_{\mathsf{q}}$ directly into the local parameter update, which is now written
\begin{subequations}
    \label{eq:alg-ensrf-hml-local-param-update-tap}
    \begin{align}
        \Delta\bar{\mathbf{q}} &= \zeta_{\mathsf{q}}\mathbf{B}_{\mathsf{qx}}\mathbf{u}_{\mathsf{x}}, \\
        \Delta\mathbf{Z}_{\mathsf{q}} &= \zeta_{\mathsf{q}}\mathbf{B}_{\mathsf{qx}}\mathbf{U}_{\mathsf{x}}.
    \end{align}
\end{subequations}
Hereafter, $\zeta_{\mathsf{q}}$ is called the \emph{local tapering} parameter, not to be confused with the (global) tapering parameter $\zeta_{\mathsf{p}}$\footnote{For completeness, we mention that, once again, it is possible to use a vector algorithmic parameter $\boldsymbol{\zeta}_{\mathsf{q}}$ instead of the scalar algorithmic parameter $\zeta_{\mathsf{q}}$, but we chose not to for the sake of simplicity.}.

To conclude, the LEnSRF-HML analysis is summarised in \cref{alg:alg-ensrf-hml}. By construction, it is equivalent to the generic LEnSRF analysis described in \cref{sssec:alg-generic-ensrf-analysis}. In the limit where localisation is disabled ($\boldsymbol{\rho}=\boldsymbol{\Pi}$, the matrix full of ones), the EnSRF-HML analysis is retrieved, and is equivalent to the generic EnSRF analysis described in \cref{sssec:alg-generic-ensrf-analysis}. Note that this algorithm explicitly uses the tangent linear operator $\mathbf{H}_{\mathsf{x}}$ of $\boldsymbol{\mathcal{H}}_{\mathsf{x}}$, which may be mandatory to assimilate non-local observations. However, if the observations are local, it is possible to derive an alternative algorithm which does not explicitly use $\mathbf{H}_{\mathsf{x}}$, but is nonetheless equivalent to the original algorithm for local, linear observation operators. This algorithm is given in \cref{sec:variant-EnSRF-HML}.

\begin{algorithm}
\caption{LEnSRF-HML analysis}
\label{alg:alg-ensrf-hml}
\begin{algorithmic}[1]
\renewcommand{\algorithmicrequire}{\textbf{Parameters:}}
\renewcommand{\algorithmicensure}{\textbf{Input:}}
\renewcommand{\algorithmiccomment}[1]{\hfill$\triangleright$~\textit{#1}}
\REQUIRE{localisation matrices $\boldsymbol{\rho}_{\mathsf{xx}}$ and $\boldsymbol{\rho}_{\mathsf{qx}}$, tapering parameters $\zeta_{\mathsf{p}}$ and $\zeta_{\mathsf{q}}$}
\ENSURE{Forecast ensemble $\mathbf{E}^{\mathsf{f}}$}
\STATE{$\bar{\mathbf{z}}^{\mathsf{f}}=\mathbf{E}^{\mathsf{f}}\mathbf{1}/N_{\mathsf{e}}$}
\STATE{$\mathbf{Z}^{\mathsf{f}}=\left(\mathbf{E}^{\mathsf{f}}-\bar{\mathbf{z}}^{\mathsf{f}}\mathbf{1}^{\top}\right)/\sqrt{N_{\mathsf{e}}-1}$}
\STATE{$\mathbf{Y} =  \mathbf{R}^{-1/2}\boldsymbol{\mathcal{H}}\left(\mathbf{E}^{\mathsf{f}}\right)\left(\mathbf{I} - \mathbf{1}\mathbf{1}^\top/N_\mathsf{e}\right)/\sqrt{N_\mathsf{e}-1}$}
\STATE{$\mathbf{B}_{\mathsf{xx}}=\boldsymbol{\rho}_{\mathsf{xx}}\circ\left[\mathbf{Z}^{\mathsf{f}}_{\mathsf{x}}\left(\mathbf{Z}^{\mathsf{f}}_{\mathsf{x}}\right)^{\top}\right]$}
\STATE{$\mathbf{B}_{\mathsf{qx}}=\boldsymbol{\rho}_{\mathsf{qx}}\circ\left[\mathbf{Z}^{\mathsf{f}}_{\mathsf{q}}\left(\mathbf{Z}^{\mathsf{f}}_{\mathsf{x}}\right)^{\top}\right]$}
\STATE{$\mathbf{B}_{\mathsf{px}}=\mathbf{Z}^{\mathsf{f}}_{\mathsf{p}}\left(\mathbf{Z}^{\mathsf{f}}_{\mathsf{x}}\right)^{\top}$}
\STATE{$\mathbf{T}_{\mathsf{y}} = \mathbf{I+R}^{-1/2}\mathbf{H}_{\mathsf{x}}\mathbf{B}_{\mathsf{xx}}\mathbf{H}^{\top}_{\mathsf{x}}\mathbf{R}^{-1/2}$}
\STATE{$\mathbf{u}_{\mathsf{x}} = \mathbf{H}^{\top}_{\mathsf{x}}\mathbf{R}^{-1/2}\mathbf{T}_{\mathsf{y}}^{-1}\mathbf{R}^{-1/2}\left(\mathbf{y}-\boldsymbol{\mathcal{H}}_{\mathsf{x}}\left(\bar{\mathbf{x}}^{\mathsf{f}}\right)\right)$}
\STATE{$\mathbf{U}_{\mathsf{x}} = -\mathbf{H}^{\top}_{\mathsf{x}}\mathbf{R}^{-1/2}\left(\mathbf{T}_{\mathsf{y}}+\mathbf{T}_{\mathsf{y}}^{1/2}\right)^{-1}\mathbf{Y}$}
\STATE{$\Delta\bar{\mathbf{x}}=\mathbf{B}_{\mathsf{xx}}\mathbf{u}_{\mathsf{x}}$}\COMMENT{state, mean update}
\STATE{$\Delta\bar{\mathbf{q}}=\zeta_{\mathsf{q}}\mathbf{B}_{\mathsf{qx}}\mathbf{u}_{\mathsf{x}}$}\COMMENT{local parameters, mean update}
\STATE{$\Delta\bar{\mathbf{p}}=\zeta_{\mathsf{p}}\mathbf{B}_{\mathsf{px}}\mathbf{u}_{\mathsf{x}}$}\COMMENT{global parameters, mean update}
\STATE{$\Delta\mathbf{Z}_{\mathsf{x}}=\mathbf{B}_{\mathsf{xx}}\mathbf{U}_{\mathsf{x}}$}\COMMENT{state, perturbation update}
\STATE{$\Delta\mathbf{Z}_{\mathsf{q}}=\zeta_{\mathsf{q}}\mathbf{B}_{\mathsf{qx}}\mathbf{U}_{\mathsf{x}}$}\COMMENT{local parameters, perturbation update}
\STATE{$\Delta\mathbf{Z}_{\mathsf{p}}=\zeta_{\mathsf{p}}\mathbf{B}_{\mathsf{px}}\mathbf{U}_{\mathsf{x}}$}\COMMENT{global parameters, perturbation update}
\RETURN{$\mathbf{E}^{\mathsf{a}}=\left(\bar{\mathbf{z}}^{\mathsf{f}}+\Delta\bar{\mathbf{z}}\right)\mathbf{1}^{\top}+\sqrt{N_{\mathsf{e}}-1}\left(\mathbf{Z}^{\mathsf{f}}+\Delta\mathbf{Z}\right)$}\COMMENT{analysis ensemble}
\end{algorithmic}
\end{algorithm}

\subsection{The ensemble transform Kalman filter}
\label{ssec:alg-etkf}

We now focus on the EnKFs with DL, for which the LETKF is exemplar. 

\subsubsection{The generic (local) ETKF analysis}
\label{sssec:alg-generic-etkf-analysis}

The generic ETKF analysis is given by the following set of equations:
\begin{subequations}
    \label{eq:alg-generic-etkf-update}
    \begin{align}
        \label{eq:alg-generic-etkf-mean-update}
        \Delta\bar{\mathbf{z}} &= \mathbf{Z}^{\mathsf{f}}\mathbf{w}^{\mathsf{a}}, \\
        \label{eq:alg-generic-etkf-pert-update}
        \Delta\mathbf{Z} &= \mathbf{Z}^{\mathsf{f}}\left(\mathbf{T}_{\mathsf{e}}^{-1/2}-\mathbf{I}\right).
    \end{align}
\end{subequations}
\Cref{eq:alg-generic-etkf-mean-update} is known as the mean update and \cref{eq:alg-generic-etkf-pert-update} as the perturbation update. In these equations, $\mathbf{Y}\in\mathbb{R}^{N_{\mathsf{y}}\times N_{\mathsf{e}}}$, $\mathbf{T}_{\mathsf{e}}\in\mathbb{R}^{N_{\mathsf{e}}\times N_{\mathsf{e}}}$, and 
$\mathbf{w}^{\mathsf{a}}\in\mathbb{R}^{N_{\mathsf{e}}}$ are ancillary variables defined by
\begin{subequations}
    \label{eq:alg-generic-etkf-def-ancillary}
    \begin{align}
        \mathbf{Y} &\triangleq \mathbf{R}^{-1/2}\boldsymbol{\mathcal{H}}\left(\mathbf{Z}^{\mathsf{f}}\right), \\
        \mathbf{T}_{\mathsf{e}} &\triangleq \mathbf{I+Y}^{\top}\mathbf{Y}, \\
        \mathbf{w}^{\mathsf{a}} &\triangleq \mathbf{T}_{\mathsf{e}}^{-1}\mathbf{Y}^{\top}\mathbf{R}^{-1/2}\left(\mathbf{y}-\boldsymbol{\mathcal{H}}\left(\bar{\mathbf{z}}^{\mathsf{f}}\right)\right).
    \end{align}
\end{subequations}
Note that this definition of $\mathbf{Y}$ is consistent with the definition in line $3$ of \cref{alg:alg-ensrf-hml}.

The main advantage of the ETKF analysis is that the linear algebra is expressed in the ensemble space ($\mathbb{R}^{N_{\mathsf{e}}}$), which is usually much smaller than both the number of observations and the augmented state space dimension ($N_{\mathsf{e}}\ll N_{\mathsf{y}},N_{\mathsf{z}}$). Unfortunately, CL expressed in the augmented state space cannot be used in the analysis. Nevertheless, DL can be included in the ETKF by making the analysis local following \citet{hunt2007,nerger2007}. 

For each augmented state variable $n\in\left\{1, \ldots, N_{\mathsf{z}}\right\}$, the inverse of the observation error covariance is tapered:
\begin{equation}
    \mathbf{R}^{-1}_{n} \triangleq \boldsymbol{\rho}_{n} \circ \mathbf{R}^{-1},
\end{equation}
where $\boldsymbol{\rho}_{n}\in\mathbb{R}^{N_{\mathsf{y}}\times N_{\mathsf{y}}}$ is the localisation matrix in observation space for the $n$-th variable, a correlation matrix which depends on the geometry of the observations relative to the $n$-th variable. This yields local variants of $\mathbf{T}_{\mathsf{e}}$ and $\mathbf{w}^{\mathsf{a}}$ which are used to compute the $n$-th row of the mean and perturbation updates $\Delta\bar{\mathbf{z}}$ and $\Delta\mathbf{Z}$. This describes the LETKF analysis. By construction, the localisation matrix $\boldsymbol{\rho}_{n}$ is rigorously defined only when both the observations and the $n$-th variable are local.

A key asset of the ETKF is that \cref{eq:alg-generic-etkf-update}, which describes the generic ETKF analysis, can also be used to implement the ETKF-ML analysis in a very efficient way. By contrast, the generic LETKF analysis described above cannot be used to implement the LETKF-ML analysis because the localisation matrix $\boldsymbol{\rho}_{n}$ cannot be rigorously defined for global model parameters.

Therefore, in the following sections, we derive an equivalent update for the ETKF-ML analysis. The goal is to provide an update scheme equivalent to \cref{eq:alg-generic-etkf-update} in the global case while being generalisable to DL with global model parameters. In \cref{sssec:alg-etkf-ml-analysis}, we only consider global model parameters, repeating but improving upon \citet{bocquet2021}, and in \cref{sssec:alg-etkf-hml-analysis}, we consider the general case with both global and local model parameters.

\subsubsection{The (local) ETKF-ML analysis}
\label{sssec:alg-etkf-ml-analysis}

Let us start with global parameters only (\textit{i.e.} $N_{\mathsf{q}}=0$). Following \citet{bocquet2021}, it is possible to separate state and parameter update in the analysis. The state update is performed using the same ensemble transform as in the generic LETKF:
\begin{subequations}
    \label{eq:alg-etkf-ml-state-update}
    \begin{align}
        \Delta\bar{\mathbf{x}} &= \mathbf{Z}^{\mathsf{f}}_{\mathsf{x}}\mathbf{w}^{\mathsf{a}}, \\
        \Delta\mathbf{Z}_{\mathsf{x}} &= \mathbf{Z}^{\mathsf{f}}_{\mathsf{x}}\left(\mathbf{T}_{\mathsf{e}}^{-1/2}-\mathbf{I}\right),
    \end{align}
\end{subequations}
and the parameter update is performed using the pseudo-inverse formulae
\cref{eq:alg-ensrf-ml-param-update-original-pinv}, which we recall here:
\begin{subequations}
    \label{eq:alg-etkf-ml-param-update-original-pinv}
    \begin{align}
        \Delta\bar{\mathbf{p}} &= \mathbf{Z}^{\mathsf{f}}_{\mathsf{p}}\left(\mathbf{Z}^{\mathsf{f}}_{\mathsf{x}}\right)^{+}\Delta\bar{\mathbf{x}}, \\
        \Delta\mathbf{Z}_{\mathsf{p}} &= \mathbf{Z}^{\mathsf{f}}_{\mathsf{p}}\left(\mathbf{Z}^{\mathsf{f}}_{\mathsf{x}}\right)^{+}\Delta\mathbf{Z}_{\mathsf{x}}.
    \end{align}
\end{subequations}

When enforcing DL, the state update \cref{eq:alg-etkf-ml-state-update} is made local (following the method described in \cref{sssec:alg-generic-etkf-analysis}), while the parameter update \cref{eq:alg-etkf-ml-param-update-original-pinv} is only indirectly localised. Indeed, as explained in \cref{sssec:alg-ensrf-ml-analysis}, the parameter update is a regression of the state update \cref{eq:alg-etkf-ml-state-update}, which is localised, into the parameter subspace. This update, combined with the state update \cref{eq:alg-etkf-ml-state-update}, defines the LETKF-ML analysis as originally proposed in \citet{bocquet2021}.

However, \cref{eq:alg-ensrf-ml-pinv-equivalence} shows that without localisation, $\mathbf{Z}^{\mathsf{f}}_{\mathsf{p}}\left(\mathbf{Z}^{\mathsf{f}}_{\mathsf{x}}\right)^{+}$ is equal to $\mathbf{B}_{\mathsf{px}}\mathbf{B}^{+}_{\mathsf{xx}}$ which, with CL, becomes $\mathbf{B}_{\mathsf{px}}\mathbf{B}^{-1}_{\mathsf{xx}}$ where $\mathbf{B}_{\mathsf{xx}}$ has been localised with $\boldsymbol{\rho}_{\mathsf{xx}}$. This last localisation footprint is missing in the original LETKF-ML analysis in \citet{bocquet2021}; we have numerically checked that, although working as expected, it makes the LETKF-ML distinctively not as accurate as the LEnSRF-ML algorithm. To fix this issue, we propose a more consistent approach for the parameter update.

Instead of using the pseudo-inverse formulae \cref{eq:alg-etkf-ml-param-update-original-pinv}, we propose to use the parameter update of the EnSRF-ML analysis, namely \cref{eq:alg-ensrf-ml-param-mean-update,eq:alg-ensrf-ml-param-pert-update}. With the additional assumption $\mathbf{H}_{\mathsf{x}}\mathbf{Z}^{\mathsf{f}}_{\mathsf{x}}=\mathbf{H}\mathbf{Z}^{\mathsf{f}}\approx\boldsymbol{\mathcal{H}}\left(\mathbf{Z}^{\mathsf{f}}\right)$, this update can be re-written as follows:
\begin{subequations}
    \label{eq:alg-etkf-ml-param-update}
    \begin{align}
        \Delta\bar{\mathbf{p}} &= \mathbf{Z}^{\mathsf{f}}_{\mathsf{p}}\mathbf{Y}^{\top}\mathbf{u}_{\mathsf{y}}, \label{eq:alg-etkf-ml-param-update-mean}\\
        \Delta\mathbf{Z}_{\mathsf{p}} &= \mathbf{Z}^{\mathsf{f}}_{\mathsf{p}}\mathbf{Y}^{\top}\mathbf{U}_{\mathsf{y}},\label{eq:alg-etkf-ml-param-update-pert}
    \end{align}
\end{subequations}
where the uncorrelated increments $\mathbf{u}_{\mathsf{y}}$ and $\mathbf{U}_{\mathsf{y}}$ are the counterparts of $\mathbf{u}_{\mathsf{x}}$ and $\mathbf{U}_{\mathsf{x}}$ in observation space, given by
\begin{subequations}
    \label{eq:alg-etkf-ml-def-u}
    \begin{align}
        \label{eq:alg-etkf-ml-def-u-mean}
        \mathbf{u}_{\mathsf{y}} &= \mathbf{R}^{-1/2}\left(\mathbf{y}-\boldsymbol{\mathcal{H}}\left(\bar{\mathbf{z}}^{\mathsf{f}}\right)\right)-\mathbf{Yw}^{\mathsf{a}}, \\
        \label{eq:alg-etkf-ml-def-u-pert}
        \mathbf{U}_{\mathsf{y}} &= -\mathbf{Y}\left(\mathbf{T}_{\mathsf{e}}+\mathbf{T}_{\mathsf{e}}^{1/2}\right)^{-1}.
    \end{align}
\end{subequations}
This parameter update, combined with the state update \cref{eq:alg-etkf-ml-state-update}, defines the ETKF-ML analysis used in this paper. A proof of these formulae can be found in \cref{app:proof-corrected-etkf-ml-formulae}.

Enforcing DL in the parameter update of this new ETKF-ML analysis is straightforward. First, the construction of the uncorrelated increments $\mathbf{u}_{\mathsf{y}}$ and $\mathbf{U}_{\mathsf{y}}$ with \cref{eq:alg-etkf-ml-def-u} is made local (following the method described in \cref{sssec:alg-generic-etkf-analysis}), and then the parameter update is computed (globally) with \cref{eq:alg-etkf-ml-param-update}. The resulting LETKF-ML analysis has exactly the same amount of localisation footprints as the LEnSRF-ML analysis. It theoretically improves upon the approximate technique proposed in \citet{aksoy2006} as it makes the update rigorous. 

Finally, as proposed in \citet{bocquet2021} and taking again inspiration from the EnSRF-ML analysis, it is possible to taper the parameter update and hence to replace \cref{eq:alg-etkf-ml-param-update} by
\begin{subequations}
    \label{eq:alg-etkf-ml-param-update-taper}
    \begin{align}
        \Delta\bar{\mathbf{p}} &= \zeta_{\mathsf{p}}\mathbf{Z}^{\mathsf{f}}_{\mathsf{p}}\mathbf{Y}^{\top}\mathbf{u}_{\mathsf{y}}, \\
        \Delta\mathbf{Z}_{\mathsf{p}} &= \zeta_{\mathsf{p}}\mathbf{Z}^{\mathsf{f}}_{\mathsf{p}}\mathbf{Y}^{\top}\mathbf{U}_{\mathsf{y}},
    \end{align}
\end{subequations}
where $\zeta_{\mathsf{p}}$ is the global tapering parameter. In the EnSRF-ML analysis, the values of $\zeta_{\mathsf{p}}$ are bounded by the fact that they are used in the definition of a positive definite matrix. By contrast here, there is no such constraint and $\zeta_{\mathsf{p}}$ can take arbitrary values.

\subsubsection{The (local) ETKF-HML analysis}
\label{sssec:alg-etkf-hml-analysis}

We now extend the ETKF-ML analysis to the case where we have both global and local parameters to estimate. For this problem, we keep the state update and the global parameter update of the ETKF-ML analysis, namely \cref{eq:alg-etkf-ml-state-update,eq:alg-etkf-ml-param-update-taper}, and we need to provide an update for the local parameters.

We choose to perform the local parameter update using the same ensemble transform as in the generic ETKF, with the addition of the local tapering parameter:
\begin{subequations}
    \label{eq:alg-etkf-hml-local-param-update}
    \begin{align}
        \Delta\bar{\mathbf{q}} &= \zeta_{\mathsf{q}}\mathbf{Z}^{\mathsf{f}}_{\mathsf{q}}\mathbf{w}^{\mathsf{a}}, \\
        \Delta\mathbf{Z}_{\mathsf{q}} &= \zeta_{\mathsf{q}}\mathbf{Z}^{\mathsf{f}}_{\mathsf{q}}\left(\mathbf{T}_{\mathsf{e}}^{-1/2}-\mathbf{I}\right).
    \end{align}
\end{subequations}
This update, combined with the state update \cref{eq:alg-etkf-ml-state-update} and the global parameter update \cref{eq:alg-etkf-ml-param-update-taper}, defines the ETKF-HML analysis. The local tapering parameter $\zeta_{\mathsf{q}}$ enables a full similarity between the ETKF-HML and EnSRF-HML analyses.

Enforcing DL in this ETKF-HML is straightforward: the local parameter update  is made local following the method described in \cref{sssec:alg-generic-etkf-analysis}. Note however, that we do not make the assumption that the local parameters and the state variables follow the same geometry. Therefore, a rigorous definition of the LETKF-HML analysis could require two sets of localisation matrices: $\left\{\boldsymbol{\rho}^{\mathsf{x}}_{n}, n=1, \ldots, N_{\mathsf{x}}\right\}$ for the state variables and $\left\{\boldsymbol{\rho}^{\mathsf{q}}_{m}, m=1, \ldots, N_{\mathsf{q}}\right\}$ for the local parameters. Hence the local state updates and local parameter updates are computed in two different localisation loops. If the geometry of the local parameters coincides with that of the state variables (\textit{i.e.} if the local parameters and the state variables are co-located), then the two localisation loops can potentially be merged.

To conclude, the LETKF-HML analysis is summarised in \cref{alg:alg-etkf-hml}. In this algorithm, we make the assumption that the observation operator is \textit{fully} local. Specifically, we hypothesise the existence of a map $h:p\mapsto h\left(p\right)$ from $[1\ldots N_{\mathsf{y}}]$ to $[1\ldots N_{\mathsf{x}}]$, which, to each index $p$ of any observation $[\mathbf{y}]_p$ associates the index $n=h(p)$ of the grid cell where the observation belongs and of which it is representative. Hence, for any $p \in [1\ldots N_{\mathsf{y}}]$, the $p$-th observation can be written $[\mathbf{y}]_p = \mathcal{H}_{\mathsf{x}, p}\left([\mathbf{x}]_{h(p)}\right)$. If needed, the algorithm can be generalised to more complex observation operators, provided that they are local, typically interpolation operators.

In the limit where localisation is disabled (for all $n\in\left\{1, \ldots, N_{\mathsf{x}}\right\}$, and $m\in\left\{1, \ldots, N_{\mathsf{x}}\right\}$, $\boldsymbol{\rho}^{\mathsf{x}}_{n}=\boldsymbol{\rho}^{\mathsf{q}}_{m}=\boldsymbol{\Pi}$, the matrix full of ones), one recovers the ETKF-HML analysis, which is equivalent to the generic ETKF analysis described in \cref{sssec:alg-generic-etkf-analysis}. Furthermore, the generic ETKF analysis being equivalent to the generic EnSRF analysis, we conclude that the ETKF-HML analysis is equivalent to the EnSRF-HML analysis. However, even though the LEnSRF-HML analysis is equivalent to the generic LEnSRF analysis, the LETKF-HML analysis is not equivalent to the generic LETKF analysis which is not defined (because of the global parameters). Finally, the parameter localisation is somewhat similar between the LETKF-HML and the LEnSRF-HML analyses, which is why we expect the difference in performance between the LETKF-HML and the LEnSRF-HML algorithms to be of the same order as the difference in performance between the LETKF and the LEnSRF algorithms \citep{sakov2011}.

The algorithms presented in \cref{ssec:alg-ensrf} and \cref{ssec:alg-etkf} are summarised in \cref{tab:enkf-ml-family} of \cref{sec:resume-algo}.

\begin{algorithm}
\caption{LETKF-HML analysis for a fully local observation operator}
\label{alg:alg-etkf-hml}
\begin{algorithmic}[1]
\renewcommand{\algorithmicrequire}{\textbf{Parameters:}}
\renewcommand{\algorithmicensure}{\textbf{Input:}}
\renewcommand{\algorithmiccomment}[1]{\hfill$\triangleright$~\textit{#1}}
\REQUIRE{localisation matrices $\left\{\boldsymbol{\rho}^{\mathsf{x}}_{n}, n=1, \ldots, N_{\mathsf{x}}\right\}$ and $\left\{\boldsymbol{\rho}^{\mathsf{q}}_{m}, m=1, \ldots, N_{\mathsf{q}}\right\}$, tapering parameters $\zeta_{\mathsf{p}}$ and $\zeta_{\mathsf{q}}$}
\ENSURE{Forecast ensemble $\mathbf{E}^{\mathsf{f}}$}
\STATE{$\bar{\mathbf{z}}^{\mathsf{f}}=\mathbf{E}^{\mathsf{f}}\mathbf{1}/N_{\mathsf{e}}$}
\STATE{$\mathbf{Z}^{\mathsf{f}}=\left(\mathbf{E}^{\mathsf{f}}-\bar{\mathbf{z}}^{\mathsf{f}}\mathbf{1}^{\top}\right)/\sqrt{N_{\mathsf{e}}-1}$}
\STATE{$\mathbf{Y} =  \mathbf{R}^{-1/2}\boldsymbol{\mathcal{H}}\left(\mathbf{E}^{\mathsf{f}}\right)\left(\mathbf{I} - \mathbf{1}\mathbf{1}^\top/N_\mathsf{e}\right)/\sqrt{N_\mathsf{e}-1}$}
\FOR{$n=1$ \TO $N_{\mathsf{x}}$}
  \STATE{$\mathbf{Y}_n = \boldsymbol{\rho}^\mathsf{x}_{n}\circ\mathbf{Y}$}
    \STATE{$\mathbf{T}_{n}=\mathbf{I}+\mathbf{Y}_n^{\top}\mathbf{Y}_n$}
    \STATE{$\boldsymbol{\delta}_n=\boldsymbol{\rho}^\mathsf{x}_{n}\circ\mathbf{R}^{-1/2}\left(\mathbf{y}-\boldsymbol{\mathcal{H}}_{\mathsf{x}}\left(\bar{\mathbf{x}}^{\mathsf{f}}\right)\right)$}
    \STATE{$\mathbf{w}^{\mathsf{a}}_{n}=\mathbf{T}_{n}^{-1}\mathbf{Y}^{\top}_n\boldsymbol{\delta}_n$}
    \FOR{$p\in h^{-1}(n)$}
    \STATE{$\left[\mathbf{u}_{\mathsf{y}}\right]_{p}=\left[\boldsymbol{\delta}_n-\mathbf{Y}_n \mathbf{w}^{\mathsf{a}}_{n}\right]_{p}$}
    \STATE{$\left[\mathbf{U}_{\mathsf{y}}\right]_{p, :} = -\left[\mathbf{Y}_n\left(\mathbf{T}_{n}+\mathbf{T}_{n}^{1/2}\right)^{-1}\right]_{p, :}$}
    \ENDFOR
    \STATE{$\left[\Delta\bar{\mathbf{x}}\right]_{n}=\left[\mathbf{Z}^{\mathsf{f}}_{\mathsf{x}}\mathbf{w}^{\mathsf{a}}_{n}\right]_{n}$}\COMMENT{state, mean update [local]}
    \STATE{$\left[\Delta\mathbf{Z}_{\mathsf{x}}\right]_{n, :}=\left[\mathbf{Z}^{\mathsf{f}}_{\mathsf{x}}\left(\mathbf{T}^{-1/2}_{n}-\mathbf{I}\right)\right]_{n, :}$}\COMMENT{state, perturbation update [local]}
\ENDFOR
\FOR{$m=1$ \TO $N_{\mathsf{q}}$}
 \STATE{$\mathbf{Y}_m = \boldsymbol{\rho}^\mathsf{x}_{m}\circ\mathbf{Y}$}
    \STATE{$\mathbf{T}_{m}=\mathbf{I}+\mathbf{Y}_m^{\top}\mathbf{Y}_m$}
    \STATE{$\boldsymbol{\delta}_m=\mathbf{R}^{-1/2}\left(\mathbf{y}-\boldsymbol{\mathcal{H}}_{\mathsf{x}}\left(\bar{\mathbf{x}}^{\mathsf{f}}\right)\right)$}
    \STATE{$\mathbf{w}^{\mathsf{a}}_{m}=\mathbf{T}_{m}^{-1}\mathbf{Y}^{\top}_m\boldsymbol{\delta}_m$}
    \STATE{$\left[\Delta\bar{\mathbf{q}}\right]_{m}=\zeta_{\mathsf{q}}\left[\mathbf{Z}^{\mathsf{f}}_{\mathsf{q}}\mathbf{w}^{\mathsf{a}}_{m}\right]_{m}$}\COMMENT{local parameters, mean update [local]}
    \STATE{$\left[\Delta\mathbf{Z}_{\mathsf{q}}\right]_{m, :}=\zeta_{\mathsf{q}}\left[\mathbf{Z}^{\mathsf{f}}_{\mathsf{q}}\left(\mathbf{T}^{-1/2}_{m}-\mathbf{I}\right)\right]_{m, :}$}\COMMENT{local parameters, perturbation update [local]}
\ENDFOR
\STATE{$\Delta\bar{\mathbf{p}} = \zeta_{\mathsf{p}}\mathbf{Z}^{\mathsf{f}}_{\mathsf{p}}\mathbf{Y}^{\top}\mathbf{u}_{\mathsf{y}}$}\COMMENT{global parameters, mean update}
\STATE{$\Delta\mathbf{Z}_{\mathsf{p}} = \zeta_{\mathsf{p}}\mathbf{Z}^{\mathsf{f}}_{\mathsf{p}}\mathbf{Y}^{\top}\mathbf{U}_{\mathsf{y}}$}\COMMENT{global parameters, perturbation update}
\RETURN{$\mathbf{E}^{\mathsf{a}}=\left(\bar{\mathbf{z}}^{\mathsf{f}}+\Delta\bar{\mathbf{z}}\right)\mathbf{1}^{\top}+\sqrt{N_{\mathsf{e}}-1}\left(\mathbf{Z}^{\mathsf{f}}+\Delta\mathbf{Z}\right)$}\COMMENT{analysis ensemble}
\end{algorithmic}
\end{algorithm}

%--------------------------------------------------

\section{Illustration of the EnKF-ML algorithms with a 1D model}
\label{sec:exp}

In this section, the EnKF-HML family of algorithms is first illustrated numerically using the Lorenz 1996 (L96) model \citep{lorenz1998}. The standard L96 model with 40 variables is widely used in DA to test new methods, but we choose here to use an inhomogeneous variant to illustrate the need for local parameters.

\subsection{The inhomogeneous Lorenz 1996 model}
\label{sec:per}
The L96 model is defined by a set of ODEs over a periodic domain with $N_{\mathsf{x}}$ variables, indexed by $n = 1, \dots,  N_{\mathsf{x}}$:
\begin{equation}
    \frac{\mathrm{d}x_n}{\mathrm{d}t} = (x_{n+1} - x_{n-2})x_{n-1} - x_n + F,
\end{equation}
where $F$ is the forcing coefficient and $x_1 = x_{N_{\mathsf{x}}+1}$, $x_{0}=x_{N_{\mathsf{x}}}$, and $x_{-1}=x_{N_{\mathsf{x}}-1}$ to ensure periodicity. The inhomogeneous L96 (L96i) model is a variant of the L96 model in which the constant forcing $F$ is replaced by a local forcing $F_{n}$ which depends on the state variable index $n$.

The standard L96 model uses $N_{\mathsf{x}}=40$ variables and $F=8$. For our experiments we use the L96i model with $N_{\mathsf{x}}=40$ variables as well and the local forcing is defined as
\begin{equation}
    \label{eq:sec-exp-def-local-forcing}
    F_{n} \triangleq 8 + \cos\left(\frac{2\pi n}{N_{\mathsf{x}}}\right).
\end{equation}
The model is integrated using a fourth-order Runge--Kutta scheme with a time step of $\delta t=0.05$. We checked that it has $13$ positive Lyapunov exponents and a neutral one, yielding an unstable-neutral subspace of dimension $14$.

\subsection{The surrogate model}
\label{ssec:illustration-1d-sur-model}

As explained in the beginning of \cref{sec:alg}, the EnKF-HML algorithms do not use the true model for the forecast but a surrogate model instead, whose parameters are estimated during the analysis. Following \citet{bocquet2020, bocquet2021}, we choose to use the surrogate model designed in \citet{bocquet2019b}. In this model, the tendencies are parametrised by a set of regressors called the \textit{monomials}, and are then integrated in time to build the resolvent between two time steps. This model can in principle represent any homogeneous ODE, provided that the number of monomials (which is determined by $L$, the size of the local \textit{stencil}) is sufficient. Note that this surrogate model has been implemented using neural networks in \citet{bocquet2019b}.

In our experiments, we use a stencil of $L=2$, we replace the global forcing coefficient by local forcing coefficients, and we use a fourth-order Runge--Kutta scheme with a time step of $\delta t=\num{0.05}$ to integrate the tendencies. The surrogate model is defined on $N_\mathsf{x}=40$ state variables, in a one-to-one correspondence with those of the L96i, and has a total of $\frac{3}{2}\left(L+1\right)\times\left(L+2\right)-1+N_\mathsf{x}=17+40=57$ parameters. The first $17$ parameters correspond to linear and bilinear monomial coefficients. The other $40$ parameters correspond to the local forcing coefficients. For convenience, we introduce $\sur\left(\mathbf{a}, \mathbf{f}\right)$ as the surrogate model in which the $17$ monomial coefficients are in vector $\mathbf{a}$ and the $40$ forcing coefficients are in vector $\mathbf{f}$. The equations of the surrogate model are given in \cref{sec:surrogate-model}.

\begin{figure}[tbp]
    \centering
    \includegraphics{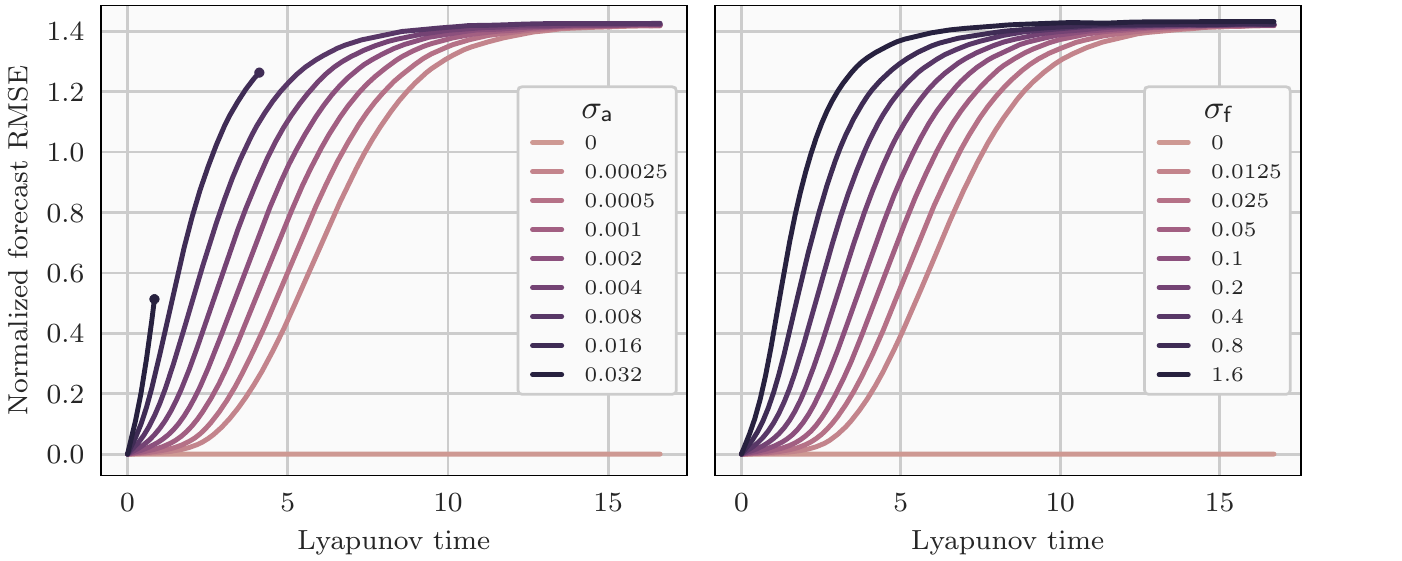}
    \caption{Forecast skill of the surrogate model $\sur\left(\mathbf{a}, \mathbf{f}\right)$ compared to $\sur\left(\mathbf{a}^{\mathsf{t}}, \mathbf{f}^{\mathsf{t}}\right)$, the true L96i model. Left panel: $\mathbf{f}=\mathbf{f}^{\mathsf{t}}$ and $\mathbf{a}=\mathbf{a}^{\mathsf{t}}+\mathbf{a}'$ with $\mathbf{a}'\sim\mathcal{N}(0, \sigma^2\mathbf{I})$ for increasing $\sigma$. Right panel: $\mathbf{a}=\mathbf{a}^{\mathsf{t}}$ and $\mathbf{f}=\mathbf{f}^{\mathsf{t}}+\mathbf{f}'$ with $\mathbf{f}'\sim\mathcal{N}(0, \sigma^2\mathbf{I})$ for increasing $\sigma$. Each experiment is repeated $\num{5000}$ times, with different parameter perturbations and different initial conditions. The curves are stopped with a dot when at least one of the $\num{5000}$ repetitions diverged. The RMSE is normalized by the variability of the true model.}
    \label{fig:surrogate_forecast}
\end{figure}

By construction, it is possible to reproduce the L96i model with a specific and unique set of parameters which we write $\mathbf{a}^{\mathsf{t}}$ and $\mathbf{f}^{\mathsf{t}}$: $\sur\left(\mathbf{a}^{\mathsf{t}}, \mathbf{f}^{\mathsf{t}}\right)$ is the L96i model. The values of $\mathbf{a}^{\mathsf{t}}$ lie in the set $\left\{-1, 0, 1\right\}$ while the values of $\mathbf{f}^{\mathsf{t}}$ are given by \cref{eq:sec-exp-def-local-forcing}. The sensitivity of the surrogate model $\sur\left(\mathbf{a}, \mathbf{f}\right)$ to $\mathbf{a}$ and $\mathbf{f}$ is illustrated in \cref{fig:surrogate_forecast} using the forecast skill, which is defined as the average integration error after a given lead time starting from the correct initial condition.

\subsection{Experimental setup}

\subsubsection{The inference problem}
\label{sssec:illustration-1d-setup-inference-pbm}

The experiments consist of twin simulations. The truth is generated using the L96i model, or equivalently using $\sur\left(\mathbf{a}^{\mathsf{t}}, \mathbf{f}^{\mathsf{t}}\right)$. The system is fully observed (the 2D system used later on is not), $\boldsymbol{\mathcal{H}}_{\mathsf{x}}\left(\mathbf{x}\right)=\mathbf{x}$, with a period of $\Delta t=\num{0.05}$, and the observations are independently perturbed with a normal distribution of error covariance matrix $\mathbf{R}=\mathbf{I}$.

Three categories of experiments are performed, with an increasing number of parameters to estimate alongside the state.
\begin{enumerate}
    \item In the first category, the goal is to estimate the $\num{17}$ monomial coefficients $\mathbf{a}$. This inference problem is very similar to the one considered in \citet{bocquet2021}. 
    \item In the second category, the goal is to estimate the $\num{40}$ forcing coefficients $\mathbf{f}$.
    \item In the third category, the goal is to estimate all $\num{57}$ coefficients.
\end{enumerate}
In all experiments, the main performance metric is the time-averaged root mean squared error (RMSE) of the state analysis. Since the set of true parameters is unique, it is also possible to compute an RMSE score for the parameter analysis. However, in such cycled experiments, we expected and we have numerically checked that small RMSE scores for the state estimation can only be obtained with accurate models, \textit{i.e.} with small RMSE scores for the parameter estimation. For this reason, we do not systematically report the parameter RMSE. Furthermore, the exact numbers of spin-up and assimilation cycles depend on the experiment and are specified later.

\subsubsection{Tested algorithms}
\label{sssec:illustration-1d-setup-tested-alg}

Our objective is to implement and test the LETKF-HML and LEnSRF-HML algorithms, for which we need to specify the set of global and local parameters $\mathbf{p}$ and $\mathbf{q}$ to be estimated alongside the state. The $\num{17}$ monomial coefficients $\mathbf{a}$ affect the model tendencies in a global way. Therefore, if they need to be estimated, they must be included in the set of global parameters $\mathbf{p}$. By contrast, the $\num{40}$ forcing coefficients $\mathbf{f}$ affect the model tendencies locally. This means that, if they need to be estimated, they can be included either in the set of global parameters $\mathbf{p}$ (\textit{i.e.}, ignoring their local nature) or in the set of local parameters $\mathbf{q}$. In order to distinguish the different algorithmic variants, we will replace the -HML suffix by a -ML suffix when there are only global parameters to estimate ($N_{\mathsf{q}}=\num{0}$) and by a -LML suffix when there are only local parameters to estimate ($N_{\mathsf{p}}=\num{0}$). This terminology is consistent with the definition of the EnKF-ML algorithms.

For comparison, we also implement and test the algorithm of \citet{aksoy2006}, hereafter called LETKF-Aksoy. This is a variant of the LETKF suited for parameter estimation, in which the global parameter update is performed through an empirical averaging of local updates. The original algorithm by \citet{aksoy2006} included a mechanism to maintain the parameter spread above a certain threshold. For simplicity, we have not used this mechanism in our experiments as we did not find it necessary. 

The setup for all the LEnKF-HML variants tested in \cref{ssec:illustration-1d-results} is summarised in \cref{tab:exp-summary}.

\begin{table}[tbp]
    %\rowcolors*{2}{}{}
    \footnotesize
    \centering
    \caption{Setup for the different algorithmic variants tested in \cref{ssec:illustration-1d-results}. For each experiment, we specify the inference problem (first column), the analysis algorithm (second column), the forecast model (third column), the definition of the set of global and local coefficients (fourth and fifth columns) and their numbers (sixth and seventh columns).}
    \label{tab:exp-summary}
    \begin{tabular}{cllcccc} \hline
        Inference problem & Algorithm & Model & $\mathbf{p}$ & $\mathbf{q}$ & $N_{\mathsf{p}}$ & $N_{\mathsf{q}}$ \\ \hline
        \multirow{3}{*}{1: $\left(\mathbf{x}, \mathbf{a}\right)$} & LEnSRF-ML & $\sur\left(\mathbf{a}, \mathbf{f}^{\mathsf{t}}\right)$ & $\mathbf{a}$ & & \num{17} & \\
        & LETKF-ML & $\sur\left(\mathbf{a}, \mathbf{f}^{\mathsf{t}}\right)$ & $\mathbf{a}$ & & \num{17} & \\
        & LETKF-Aksoy & $\sur\left(\mathbf{a}, \mathbf{f}^{\mathsf{t}}\right)$ & $\mathbf{a}$ & & \num{17} & \\\hline
        \multirow{2}{*}{2: $\left(\mathbf{x}, \mathbf{f}\right)$} & LETKF-ML & $\sur\left(\mathbf{a}^{\mathsf{t}}, \mathbf{f}\right)$ & $\mathbf{f}$ & & \num{40} & \\
        & LETKF-LML & $\sur\left(\mathbf{a}^{\mathsf{t}}, \mathbf{f}\right)$ & & $\mathbf{f}$ & & \num{40} \\ \hline
        \multirow{2}{*}{3: $\left(\mathbf{x}, \mathbf{a}, \mathbf{f}\right)$} & LETKF-HML & $\sur\left(\mathbf{a}, \mathbf{f}\right)$ & $\mathbf{a}$ & $\mathbf{f}$ & \num{17} & \num{40} \\
        & LEnSRF-HML & $\sur\left(\mathbf{a}, \mathbf{f}\right)$ & $\mathbf{a}$ & $\mathbf{f}$ & \num{17} & \num{40} \\\hline
    \end{tabular}
\end{table}

\subsubsection{Ensemble initialisation}
\label{sssec:illustration-1d-setup-ens-init}

As shown in \citet{bocquet2021}, the ensemble initialisation may have an impact on the time-averaged metric (even with a very long run). In this paper, this is less critical because we only use localised ensemble DA algorithms. Nevertheless, we stick to the initialisation method described in \citet{bocquet2021}. Namely, the $i$-th ensemble member is initialised as
\begin{equation}
    \mathbf{z}_{i} = \mathbf{z}^{\mathsf{t}} + \mathbf{z}' + \mathbf{z}''_{i}, \quad \mathbf{z}', \mathbf{z}''_{i} \sim\mathcal{N}\left(\mathbf{0}, \boldsymbol{\Sigma}\right),
\end{equation}
where $\mathbf{z}^{\mathsf{t}}$ is the true initial state, $\mathbf{z}'$ is the initial bias, and $\mathbf{z}''_{i}$ is the $i$-th asymptotically unbiased perturbation. The covariance matrix $\boldsymbol{\Sigma}$ is diagonal, equal to $\num{1}$ for the state variables and to $\num{0.2}$ for both the local and global parameters. As shown in \cref{fig:surrogate_forecast}, having a $\num{0.2}$ bias in the parameters is sufficient to make the surrogate model inaccurate.

\subsubsection{Algorithm parametrisation}

For the LEnSRF-HML analysis, \cref{alg:alg-ensrf-hml}, we need to specify two localisation matrices: the classical localisation matrix between state variables, $\boldsymbol{\rho}_{\mathsf{xx}}$, and the cross localisation matrix between state variables and local parameters $\boldsymbol{\rho}_{\mathsf{qx}}$. In all our experiments, the geometry of the local parameters (if any) is the same as the geometry of the state variables. Therefore and for the sake of simplicity, we enforce $\boldsymbol{\rho}_{\mathsf{qx}}=\boldsymbol{\rho}_{\mathsf{xx}}$ and $\boldsymbol{\rho}_{\mathsf{xx}}$ is chosen as
\begin{equation}
    \left[\boldsymbol{\rho}_{\mathsf{xx}}\right]_{mn} \triangleq \gc\left(\frac{2d\left(m, n\right)}{r}\right),
\end{equation}
where $\gc$ is the Gaspari--Cohn piecewise rational function \citep{gaspari1999},
$d\left(m, n\right)$ is the (circular) distance between the $m$-th and $n$-th variables, and $r$ is the localisation radius, the only algorithmic parameter relative to localisation. 

For the LETKF-HML analysis, \cref{alg:alg-etkf-hml}, we also need to specify two sets of localisation matrices: the classical localisation matrices between observations and state variables, $\boldsymbol{\rho}^{\mathsf{x}}_{n}$, and the localisation matrices between observations and local parameters, $\boldsymbol{\rho}^{\mathsf{q}}_{n}$. For the same reasons as above, we enforce $\boldsymbol{\rho}^{\mathsf{q}}_{n}=\boldsymbol{\rho}^{\mathsf{x}}_{n}$ and the $\boldsymbol{\rho}^{\mathsf{x}}_{n}$ matrices are chosen as
\begin{equation}
    \left[\boldsymbol{\rho}^{\mathsf{x}}_{n}\right]_{ij} \triangleq \sqrt{\gc\left(\frac{2d\left(i, n\right)}{r}\right)\gc\left(\frac{2d\left(j, n\right)}{r}\right)},
\end{equation}
where $d\left(i, n\right)$ and $d\left(j, n\right)$ are the distances between the $i$-th observation and the $n$-th variable and between the $j$-th observation and the $n$-th variable, respectively, and $r$ is the localisation radius. Besides, having the same geometry for the state variables and the local parameters means that the two for-loops in \cref{alg:alg-etkf-hml} can be merged. 

Finally, in order to mitigate the sampling errors, we use a multiplicative inflation on the prior with a uniform and constant in time coefficient $\lambda$. Preliminary experiments have shown that using different inflation coefficients for model state and model parameters does not significantly improve the scores, which is why we chose to use the same uniform inflation coefficient for all components of the augmented state. Note however that this result may not generalise to other experiments, as suggested by other studies in the literature \citep{kang2011}.

To summarise, our algorithms depend on at most four scalar parameters: the localisation radius $r$ (which parametrises the Gaspari--Cohn function), the inflation coefficient $\lambda$, and the two tapering coefficients $\zeta_{\mathsf{p}}$ and $\zeta_{\mathsf{q}}$ introduced in \cref{sec:alg}. Unless otherwise mentioned, each algorithmic parameter is optimally tuned to yield the lowest state RMSE for each experiment.

\subsection{Results}
\label{ssec:illustration-1d-results}

In this section, we present the results of our numerical experiments, organised according to the classification described in \cref{sssec:illustration-1d-setup-inference-pbm}.

\subsubsection{Estimation of the 17 monomial coefficients}
\label{sssec:illustration-1d-results-first-category}

In this first test series, the goal is to estimate the 17 monomial coefficients $\mathbf{a}$ only. As explained in \cref{sssec:illustration-1d-setup-tested-alg}, these coefficients affect the model tendencies in a global way, and hence must be included in the set of global parameters $\mathbf{p}$, which means that $N_{\mathsf{p}}=\num{17}$. For these experiments, there is no local parameter: $N_{\mathsf{q}}=\num{0}$. The setup for each LEnKF-HML variant tested in this section is recalled in \cref{tab:exp-summary} (first three rows).

There are only two minor differences between this inference problem and the one considered in \citet{bocquet2021}. First, the truth is generated using the L96i model and not the L96 model. Second, the number of parameters $N_{\mathsf{p}}$ to estimate is 17 and not 18. Indeed, the inference problem of \citet{bocquet2021} also included a global forcing coefficient. This global forcing coefficient has been replaced by the local forcing coefficients $\mathbf{f}$ (while defining the L96i model) which are not estimated in this first test series.

\begin{figure}[tbp]
\centering
\includegraphics{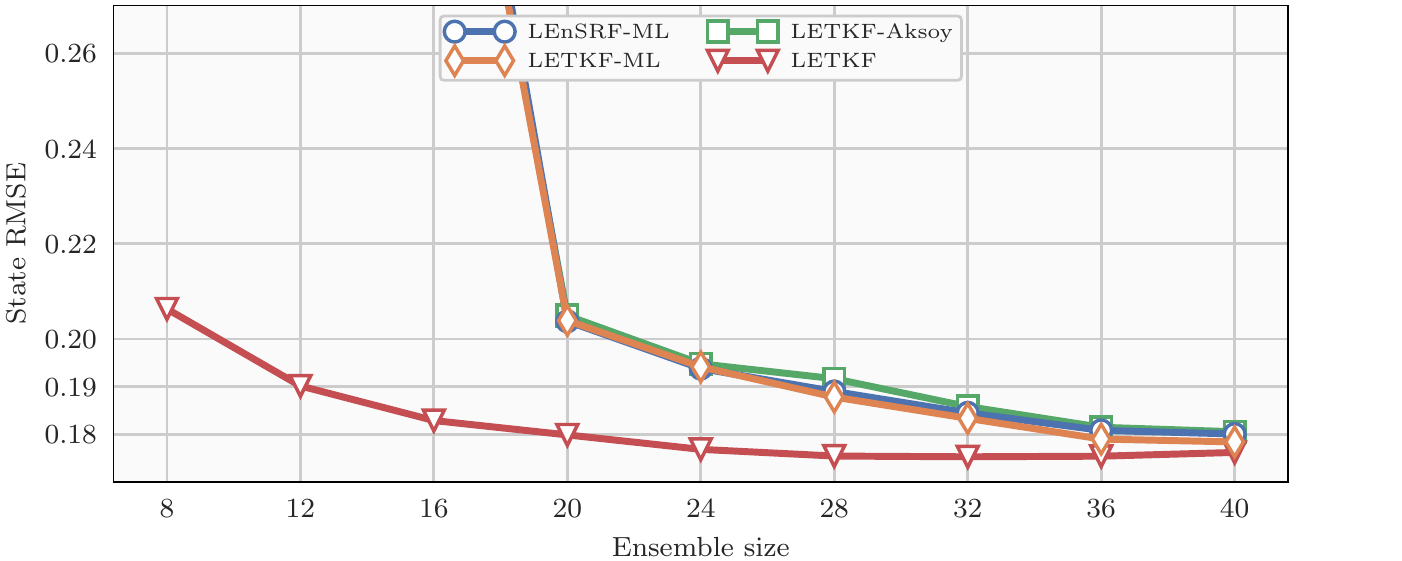}
\caption{Time-averaged state analysis RMSE as a function of the ensemble size $N_{\mathsf{e}}$ for the first test series (estimation of the 17 monomial coefficients $\mathbf{a}$) with the LEnSRF-ML (in blue), the LETKF-ML (in yellow), and the LETKF-Aksoy (in green). For reference, the red line shows the scores obtained with the LETKF when the model is known.}
\label{fig:aksoy_ml}
\end{figure}

The results are shown in \cref{fig:aksoy_ml}. The state analysis RMSE is averaged over $\num{3000}$ cycles after a spin-up period of $\num{3000}$ cycles, and over $\num{8}$ repetitions of the experiments. This is empirically sufficient to ensure the convergence of the statistical indicators.

As expected from the similarity between the inference problems, the scores obtained with the LEnSRF-ML are overall similar to those reported by \citet{bocquet2021}. Indeed, the minimal ensemble size $N_{\mathsf{e}}$ for a successful run (analysis RMSE around $\num{0.2}$) is $\num{20}$. This could be interpreted as $\num{17}$ members for the $N_{\mathsf{p}}=\num{17}$ global parameters (each global parameter is a neutral mode of the dynamics) plus a few additional members for the $N_{\mathsf{x}}=\num{40}$ state variables, for which the number of unstable and neutral modes is $\num{14}$, but which benefit from localisation.

There is almost no difference between the scores of the LEnSRF-ML and those of the LETKF-ML. This is not a surprise because the global parameter update of the LETKF-ML has been redesigned in \cref{sssec:alg-etkf-ml-analysis} to mimic that of the LEnSRF-ML, in such a way that the LEnSRF-ML and the LETKF-ML are as close to another as the LEnSRF and the LETKF.

More surprisingly, the LETKF-Aksoy method yields very similar results. Of course, the LETKF-Aksoy method makes sense: within each local domain, we obtain an estimate of the global parameters, therefore defining the global estimate as the average of the local estimates is natural. At the same time, we cannot exclude  the possibility that the global parameter estimates vary a lot over the local domains, in which case making an average may not necessarily be a good option. This is why we expected the LETKF-ML to be more robust than the LETKF-Aksoy, because the LETKF-ML provides one estimate of the global parameters consistent with all local domains, which \emph{seems} more rigorous. The similarity in scores suggest that there might be a deeper connection between the two methods. A further study is required to understand the mathematical justification of the global parameter update of the LETKF-Aksoy \citep[for example using the alternating direction method of multipliers method, see][and references therein]{boyd2011} and its potential limitations.

Finally, even though $N_{\mathsf{e}}=\num{20}$ members are sufficient for a successful run, there is still at this point a small gap between the scores of the LEnKF-ML algorithms (\textit{i.e.} with parameter estimation) and those of the LETKF (\textit{i.e.} with known model). According to \citet{bocquet2021}, this gap comes from the use of a uniform (rather than adaptive) inflation and indeed progressively vanishes as the ensemble size $N_{\mathsf{e}}$ grows.

\subsubsection{Estimation of the 40 forcing coefficients}
\label{sssec:illustration-1d-results-second-category}

In this second test series, the goal is to estimate the 40 forcing coefficients $\mathbf{f}$ only. As explained in \cref{sssec:illustration-1d-setup-tested-alg}, these coefficients affect the model tendencies in a local way, and hence they can be included either in the set of global parameters $\mathbf{p}$ or in the set of local parameters $\mathbf{q}$. Our objective is to compare the two approaches and demonstrate that parameter localisation is effective. The setup for each LEnKF-HML variant tested in this section is recalled in \cref{tab:exp-summary} (fourth and fifth rows).

\begin{figure}[tbp]
\centering
\includegraphics{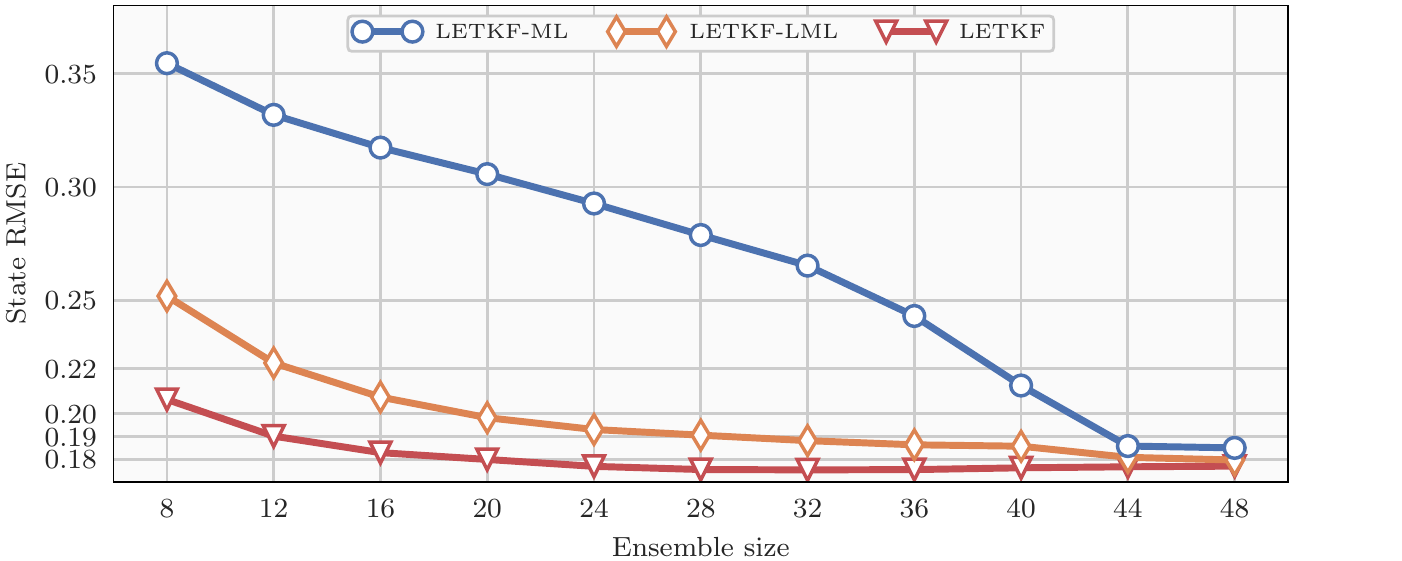}
\caption{Time-averaged state analysis RMSE as a function of the ensemble size $N_{\mathsf{e}}$ for the second test series (estimation of the 40 forcing coefficients $\mathbf{f}$) with the LETKF-ML (in blue) and the LETKF-LML (in yellow). For reference, the red line shows the scores obtained with the LETKF when the forcing coefficients are known.}
\label{fig:LETKF-ML-LML}
\end{figure}

The results are shown in \cref{fig:LETKF-ML-LML}. The state analysis RMSE is averaged over $\num{3000}$ cycles after a spin-up period of $\num{5000}$ cycles, and over $\num{8}$ repetitions of the experiments.

With the LETKF-ML, the local nature of the forcing coefficients $\mathbf{f}$ is ignored and hence the algorithm uses $N_{\mathsf{p}}=\num{40}$ global parameters and no local parameter: $N_{\mathsf{q}}=\num{0}$. As in the previous test series, we expect that the minimal ensemble size $N_{\mathsf{e}}$ for a successful run should be around $N_{\mathsf{p}}=\num{40}$ members for the global parameters (each global parameter is a neutral mode of the dynamics) plus a few additional members for the $N_{\mathsf{x}}=\num{40}$ state variables, for which the number of unstable and neutral modes is $\num{14}$, but which benefit from the localisation. This is indeed what is observed in \cref{fig:LETKF-ML-LML}. However, the divergence of the LETKF-ML for small ensembles ($N_{\mathsf{x}}\leq\num{40}$) is much less pronounced here than in the first test series. This can be explained by the fact that the surrogate model $\sur\left(\mathbf{a}, \mathbf{f}\right)$ is more sensitive to a perturbation of the monomial coefficients $\mathbf{a}$ than to a perturbation of the forcing coefficients $\mathbf{f}$, as illustrated by \cref{fig:surrogate_forecast}. In particular, the initial bias in model parameters (as described in \cref{sssec:illustration-1d-setup-ens-init}) is much weaker in relative terms in this test series than in the first one.

With the LETKF-LML, the local nature of the forcing coefficients $\mathbf{f}$ is fully exploited. Hence the algorithm uses $N_{\mathsf{q}}=\num{40}$ local parameters and no global parameter: $N_{\mathsf{p}}=\num{0}$. \cref{fig:LETKF-ML-LML} shows that the localisation of the parameters is efficient. The minimal ensemble size $N_{\mathsf{e}}$ for a successful run has been reduced from about $\num{44}$ (without the LETKF-ML) to about $\num{20}$. Furthermore, the scores obtained by the LETKF-LML are qualitatively close to those obtained by the LETKF (with known model), although there is a small gap, which corresponds to the estimation of one additional parameter per grid point.

\subsubsection{Estimation of all 57 model coefficients}
\label{sec:all_coeffs_1D}

In this third test series, the goal is to estimate the $\num{17}$ monomial coefficients $\mathbf{a}$ as well as the $\num{40}$ forcing coefficients $\mathbf{f}$. The monomial coefficients $\mathbf{a}$ must be included in the set of global parameters $\mathbf{p}$, while the forcing can be included in the set of local parameters $\mathbf{q}$. Hence, in these experiments there are $N_{\mathsf{p}}=\num{17}$ global parameters and $N_{\mathsf{q}}=\num{40}$ local parameters. The setup for each LEnKF-HML variant tested in this section is recalled in \cref{tab:exp-summary} (last two rows). 

\begin{figure}[tbp]
\centering
\includegraphics{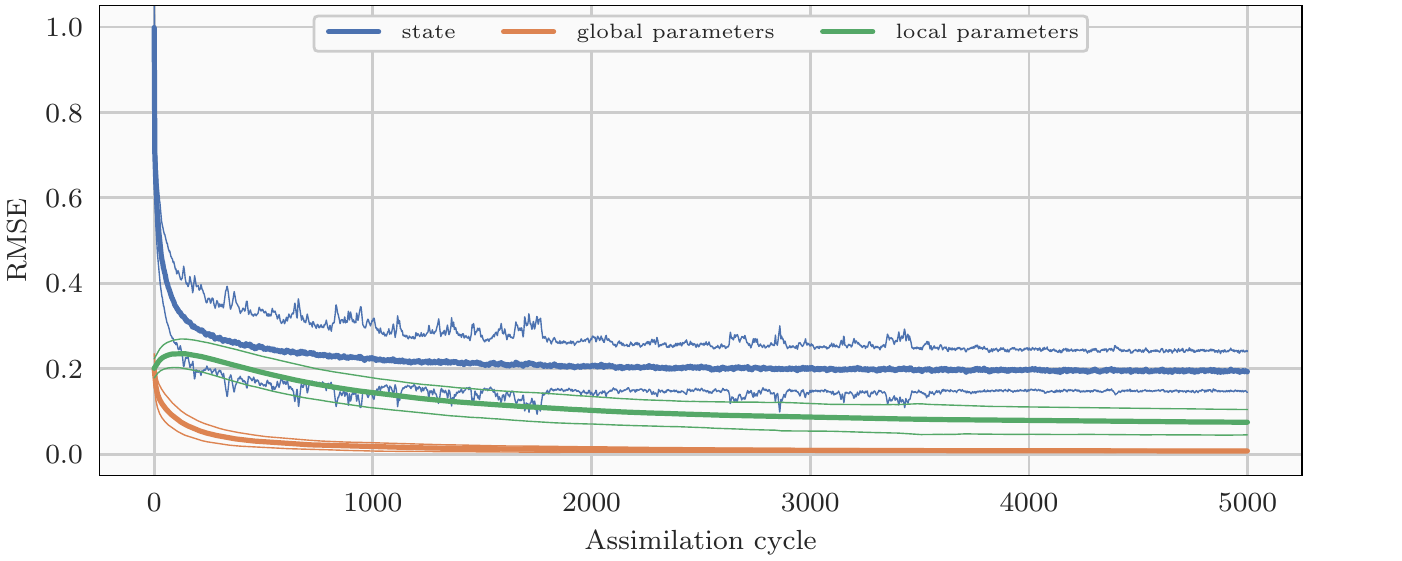}
\caption{Time series of instantaneous analysis RMSE for the third test series (estimation of all 57 model coefficients) with the LEnSRF-HML. The state RMSE is shown in blue, the global parameter RMSE in yellow, and the local parameter RMSE in green. The experiment is repeated $\num{1000}$ times, with different initial conditions and different observations. The thick line shows the average over all repetitions and the thin lines stand for the average plus or minus one standard deviation.}
\label{fig:time-LEnSRF-HML}
\end{figure}

The result of a first experiment with the LEnSRF-HML is shown in \cref{fig:time-LEnSRF-HML}. For this experiment, the ensemble size $N_{\mathsf{e}}$ is set to $\num{36}$ and the specific values for the algorithmic parameters ($r$, $\lambda$, $\zeta_{\mathsf{p}}$, and $\zeta_{\mathsf{q}}$) are chosen \emph{by trial and error}. First of all, this experiment can be qualified as successful: after a spin-up period of several thousands of cycles, the state analysis RMSE stabilises below $\num{0.2}$. Second, the improvement of the analysis is overall rather slow. Parameter estimation in ensemble DA is slow in general, but it is here most likely due to a misspecification of the algorithmic parameters. For example, increasing the inflation factor $\lambda$ could help at the beginning of the experiment, when the surrogate model is inaccurate, but would impair the analysis at the end of the experiment, when the surrogate model is more precise. Using an adaptive inflation would resolve this dilemma, but this is beyond the scope of this paper\footnote{See \citet{bocquet2021} for an efficient example of adaptive inflation scheme but in the absence of localisation.}. Third, the algorithm improves the global parameter analysis before the local parameter analysis, the state analysis RMSE seems much more correlated to the global parameter analysis RMSE than to the local parameter analysis RMSE, and the final spread of the local parameter analysis RMSE is much larger than that of the global parameter analysis RMSE. All three elements are related to the fact that the surrogate model $\sur\left(\mathbf{a}, \mathbf{f}\right)$ is more sensitive to a perturbation of the monomial coefficients $\mathbf{a}$ (which are the global parameters $\mathbf{p}$ in this experiment) than to a perturbation of the forcing coefficients $\mathbf{f}$ (which are the local parameters $\mathbf{q}$ in this experiment). Finally, the small increase in the local parameter analysis RMSE at the beginning of the experiment is once again most likely due to a misspecification of the algorithmic parameters.

\begin{figure}[tbp]
\centering
\includegraphics{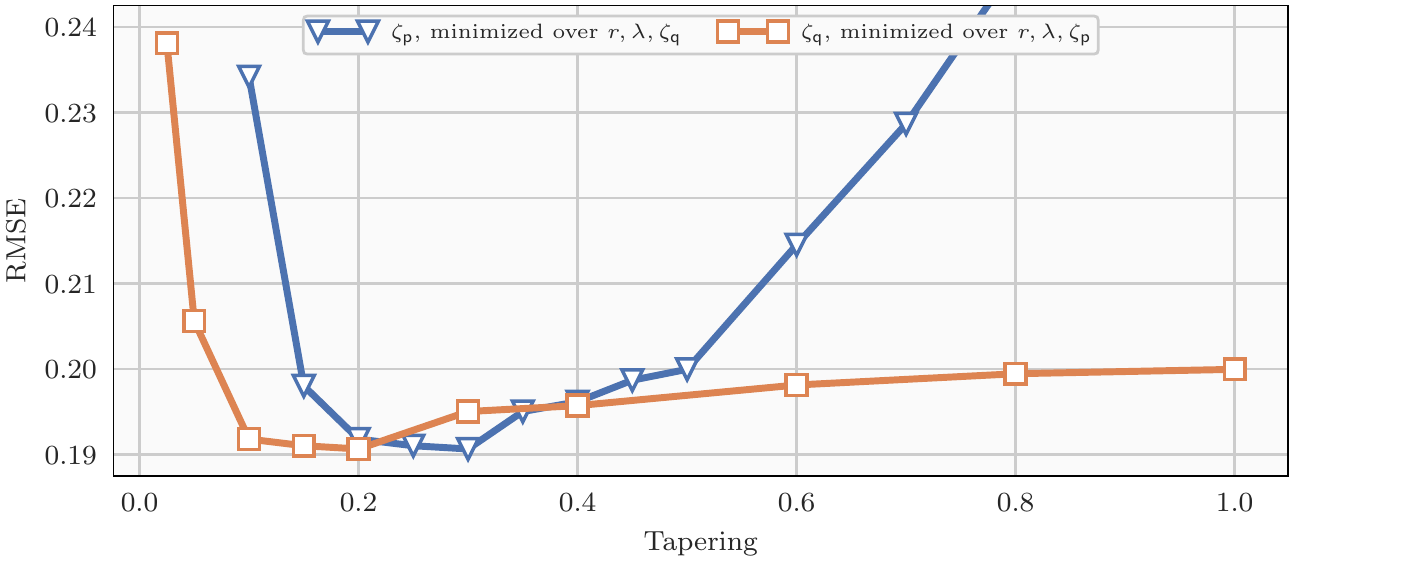}
\caption{Time-averaged state analysis RMSE as a function of the global (in blue) and local (in yellow) tapering coefficient for the third test series (estimation of all 57 model coefficients) with the LEnSRF-HML.}
\label{fig:LETKF-HML-2}
\end{figure}

After this first successful experiment, we wish to better characterise the function of each algorithmic parameter. While the role of the localisation radius $r$ and of the multiplicative inflation factor $\lambda$ are well documented in the DA literature, this is not the case for the tapering coefficients $\zeta_{\mathsf{p}}$ and $\zeta_{\mathsf{q}}$. For this reason, we show how the accuracy of the analysis depends on $\zeta_{\mathsf{p}}$ and $\zeta_{\mathsf{q}}$ in \cref{fig:LETKF-HML-2}. The state analysis RMSE is averaged over $\num{10000}$ cycles after a spin-up period of $\num{10000}$ cycles, and over $\num{8}$ repetitions of the experiments. The ensemble size $N_{\mathsf{e}}$ is kept to $\num{36}$ and in each case ($\zeta_{\mathsf{p}}$ and $\zeta_{\mathsf{q}}$), the values of the three other algorithmic parameters ($r$, $\lambda$, and the other $\zeta$) are optimally tuned to yield the lowest time-averaged state analysis RMSE. 

Let us first discuss the global tapering coefficient $\zeta_{\mathsf{p}}$. Without tapering ($\zeta_{\mathsf{p}}=\num{1}$), the algorithm fails at estimating the global parameters, and therefore the state. The global parameter update per cycle is too strong compared to the amount of information brought to the system by only one batch of observations. In a way, the algorithm is constantly overfitting the single batch of observations at each cycle. This issue is most likely due to the ensemble being too small to accurately represent the cross-correlations between state variables and global parameters, because, empirically, the need for tapering vanishes as the ensemble size grows \citep{bocquet2021}. Hence, $\zeta_{\mathsf{p}}$ can here also be seen as a \emph{relaxation} parameter. The analysis progressively improves as $\zeta_{\mathsf{p}}$ decreases, making the global parameter update slower but \emph{more robust}. Finally, the state analysis RMSE reaches an optimal value and then grows again when the tapering is too strong. Indeed, for very small values of $\zeta_{\mathsf{p}}$, the global parameter update is very slow, slow enough that the number of cycles used in the experiment, even though already large, is not enough to ensure the convergence of the statistics. Furthermore, as can be seen in \cref{fig:LETKF-HML-2}, lower values of the global tapering $\zeta_\mathsf{p}$, typically below $\num{0.1}$, yields numerical divergence of the filter since the relaxation towards a better surrogate model is too slow.

The influence of the local tapering coefficient $\zeta_{\mathsf{q}}$ is qualitatively similar to that of $\zeta_{\mathsf{p}}$ with one exception. Even if using $\zeta_{\mathsf{q}}<\num{1}$ yields better scores, tapering is not mandatory because the experiment is already successful without tapering ($\zeta_{\mathsf{q}}=\num{1}$). This is most probably due to the fact that cross-correlations between state variables and local parameters are easier to estimate thanks to parameter localisation.

\begin{figure}[htbp]
\centering
\includegraphics{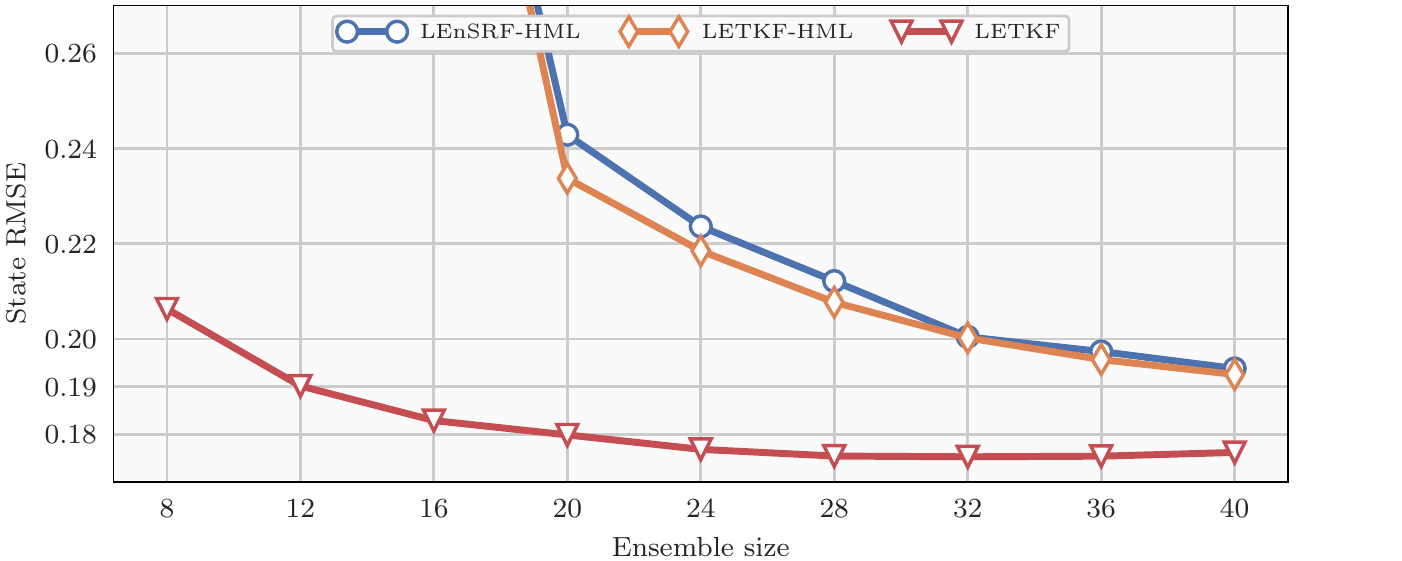}
\caption{Time-averaged state analysis RMSE as a function of the ensemble size $N_{\mathsf{e}}$ for the third test series (estimation of all $\num{57}$ model coefficients) with the LEnSRF-HML (in blue) and the LETKF-HML (in yellow). For reference, the red line shows the scores obtained with the LETKF when the model is known.}
\label{fig:LETKF-HML}
\end{figure}

Finally, we show the accuracy of the analysis as a function of the ensemble size $N_{\mathsf{e}}$ in \cref{fig:LETKF-HML}. The state analysis RMSE is averaged over $\num{10000}$ cycles after a spin-up period of $\num{10000}$ cycles, and over $\num{8}$ repetition of the experiments.

First, there is almost no difference between the scores of the LEnSRF-HML and those of the LETKF-HML in the accurate estimation part of the curves (higher $N_{\mathsf{e}}$). The similarity between both algorithms can be explained using the same argument as for the similarity between the LEnSRF-ML and the LETKF-ML in \cref{sssec:illustration-1d-results-first-category}. Second, two regimes can be qualitatively distinguished for the LEnKF-HML variants. When the ensemble size $N_{\mathsf{e}}$ is smaller than $\num{20}$, the algorithms diverge, in a way which is very similar to the divergence of the LEnKF-ML variants in \cref{sssec:illustration-1d-results-first-category}. When the ensemble size $N_{\mathsf{e}}$ is larger than $\num{20}$, the accuracy of the analysis progressively improves, in a way which is very similar to the LETKF-LML in \cref{sssec:illustration-1d-results-second-category}. These regimes can be explained as follows. In general, the LEnKF-HML estimates the most sensitive parameters first. In our experiments, the most sensitive parameters are the global parameters $\mathbf{p}$, which correspond to the $\num{17}$ monomial coefficients $\mathbf{a}$. As for the LEnKF-ML, the minimal ensemble size $N_{\mathsf{e}}$ to estimate the state and the global parameters is around 20: 17 members for the $N_{\mathsf{p}}=\num{17}$ global parameters (each global parameter is a neutral mode of the dynamics) plus a few additional members for the $N_{\mathsf{x}}=\num{40}$ state variables ($\num{14}$ unstable and neutral modes, but the assimilation is localised). However, in this third test series, using $N_{\mathsf{e}}=\num{20}$ is not sufficient because we must also estimate the $N_{\mathsf{q}}=\num{40}$ local parameters. This explains the second regime which is qualitatively similar to the LETKF-LML. In this regime, we must add $\num{12}$ additional members to decrease the analysis RMSE to $\num{0.2}$. This is less than the additional $N_{\mathsf{q}}=\num{40}$ local parameters to estimate, which shows that parameter localisation is efficient. Eventually, for larger ensembles, the scores obtained by the LEnKF-HML variants become close to those obtained by the LETKF (with known model), with a small but meaningful gap corresponding to the additional estimation of the $N_{\mathsf{p}}=\num{17}$ global parameters and of the $N_{\mathsf{q}}=\num{40}$ local parameters.

%--------------------------------------------------

\section{Two dimensional illustration with global and local parameters, covariance and domain localisations}\label{sec:2D_experiments}

In this section, we provide an illustration of a selection of EnKF-HML algorithms with the multilayer L96 (mL96) model \citep{farchi2019}, which is a two-dimensional (horizontal and vertical) extension of the standard L96 model with radiance-like (hence non-local) observations. This may seem a complicated example but it actually reflects to a large extent the requirements of a realistic, high-dimensional application of our methods.

\subsection{The multilayer Lorenz 1996 model}

The mL96 model consists in a vertical stack of $N_\mathsf{v}=\num{32}$ coupled (atmospheric) layers, each layer being a one-dimensional L96 model with $N_\mathsf{h}=\num{40}$ variables. The total state dimension is hence $N_\mathsf{x} = N_\mathsf{h}\times N_\mathsf{v}=\num{1280}$, and the model's equations are given by the following set of ODEs:
\begin{equation}
\label{eq:illustration-2d-ml96-ode}
\frac{\mathrm{d}x_{v, h}}{\mathrm{d}t} = (x_{v, h+1} - x_{v, h-2})x_{v, h-1} - x_{v, h} + F_{v, h} + \Gamma_{v+1, h} - \Gamma_{v, h},
\end{equation}
where $x_{v, h}$ is the $h$-th horizontal variable of the $v$-th vertical layer. The first terms in this equation correspond to the original L96 dynamics, where the horizontal index $h$ applies periodically in $\left\{1,\dots, N_\mathsf{h}\right\}$. The forcing term $F$ is inhomogeneous; it is set constant over each layer and decreases from $F_{1, h}=\num{8}$ for the bottom layer to $F_{N_\mathsf{v}, h}=\num{4}$ for the top layer. Finally, the last two terms correspond to the vertical coupling between adjacent layers, with
\begin{equation}
    \Gamma_{v, h} \triangleq \left\{\begin{array}{rl}
         x_{v, h} - x_{v-1, h} & \text{if } 2 \leq v \leq N_\mathsf{v}, \\
         0 & \text{otherwise}.
    \end{array}
    \right.
\end{equation}

The model is integrated using a fourth-order Runge--Kutta scheme with a time step of $\delta t=\num{0.05}$. The dimension of the unstable and neutral subspace of the dynamics is about $\num{50}$ \citep{farchi2019}.

\subsection{The surrogate model}
\label{sec:sur2D}

For this two-dimensional illustration, we use the surrogate model presented in \cref{ssec:illustration-1d-sur-model}, which we adapt in the following way.
\begin{enumerate}
    \item The $\num{17}$ monomial coefficients $\mathbf{a}$ are shared between all $N_\mathsf{v}$ layers.
    \item In theory, the number of forcing coefficients of the model is $N_\mathsf{v}\times N_\mathsf{h}$ (one for each state variable). To reduce this number and avoid an excessive initial underdetermination, we parametrise the forcing as $F_{v,h}=F_{\mathsf{v}}\left(v\right)\times F_{\mathsf{h}}\left(h\right)$, where $F_{\mathsf{v}}$ and $F_{\mathsf{h}}$ capture the vertical and horizontal variations of the forcing, respectively. The total number of forcing coefficients is hence $N_\mathsf{v}+N_\mathsf{h}$. However, to ensure the uniqueness of the decomposition, we rescale $F_{\mathsf{v}}$ and $F_{\mathsf{h}}$ in such a way that $F_{\mathsf{v}}\left(0\right)$ is always $\num{1}$. This reduces the effective number of forcing coefficients to $N_\mathsf{v}+N_\mathsf{h}-1$.
    \item In the following, the vertical coupling terms $\Gamma_{v+1, h}$ and $\Gamma_{v, h}$ are hard-coded in the model. In more advanced experiments, we have successfully learnt those parameters with success just as the rest of \cref{eq:illustration-2d-ml96-ode}, but  we do not report it for the sake of conciseness.
\end{enumerate}
As in \cref{ssec:illustration-1d-sur-model}, for convenience we introduce $\sur\left(\mathbf{a}, \mathbf{f}_{\mathsf{v}}, \mathbf{f}_{\mathsf{h}}\right)$ as the surrogate model in which the $\num{17}$ monomial coefficients are $\mathbf{a}$, the $N_\mathsf{v}-1=\num{31}$ vertical forcing coefficients are $\mathbf{f}_{\mathsf{v}}$ and the $N_\mathsf{h}=\num{40}$ horizontal forcing coefficients are $\mathbf{f}_{\mathsf{h}}$. The total number of parameters of this surrogate model is $\num{17}+\num{31}+\num{40}=\num{88}$. With this parametrisation, the true mL96 model is identifiable: by construction it can be reproduced with a given set of parameters.

\subsection{Experimental setup}
\subsubsection{The inference problem}
\label{sssec:illustration-2d-setup-inference-pbm}

The experiments consist of twin simulations. The truth is generated using the mL96 model. At each time step $\Delta t=\num{0.05}$, a total of $N_\mathsf{y} = 8\times 40 = \num{320}$ observations are generated, whose characteristics will be described in the next section.

In addition to estimating the state variables ($40\times N_\mathsf{v}=1280$ scalars), the goal is to estimate the $17$ monomial coefficients $\mathbf{a}$, the $\num{40}$ horizontal forcings $\mathbf{f}_\mathsf{h}$ and the $\num{31}$ vertical forcings in $\mathbf{f}_\mathsf{v}$. In all experiments, the main performance metric is the time-averaged RMSE of the state analysis, as, in this context with very few parameters, a small state RMSE only can be obtained with successful parameter estimation.

\subsubsection{Observation setup}

\begin{figure}[tbp]
    \centering
    \includegraphics{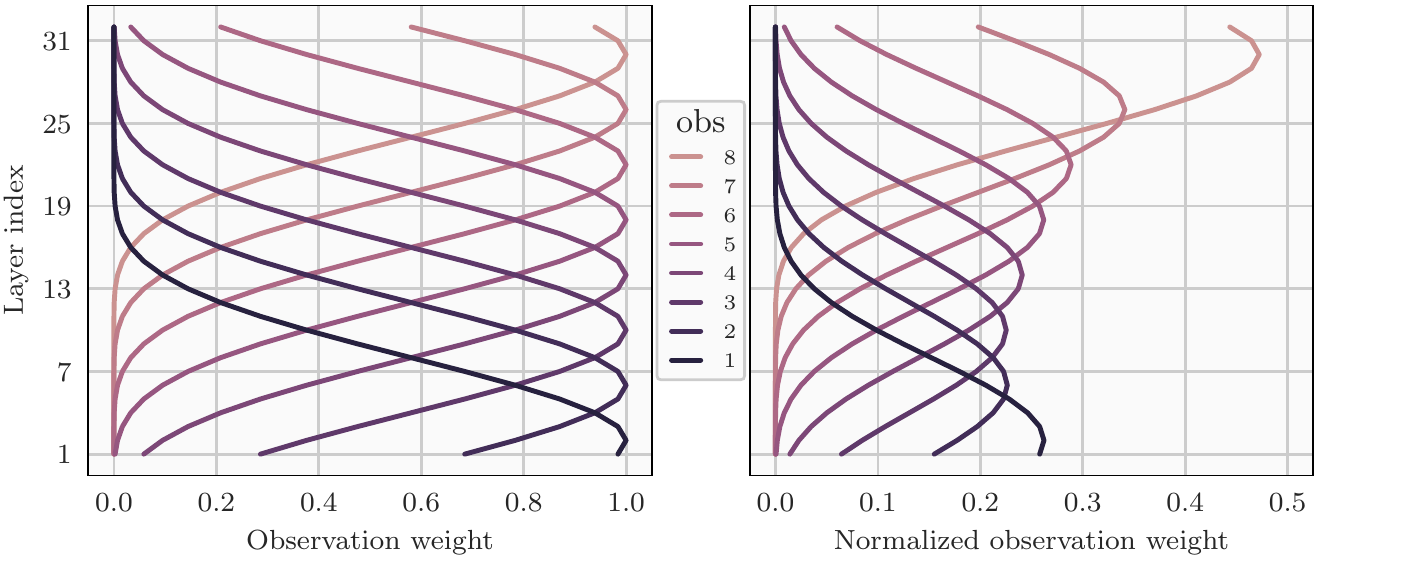}
    \caption{Averaging kernel for each of the $\num{8}$ satellite channels without (left panel) and with (right panel) normalisation.}
    \label{fig:observation}
\end{figure}

For this multilayer model, the observations consist of satellite soundings with $\num{8}$ channels. Each channel is characterised by a vertical distribution of observation weights, also called averaging kernel, which is applied to all $N_{\mathsf{h}}=\num{40}$ columns of state variables\footnote{A column is defined here as the set of $N_{\mathsf{v}}=\num{32}$ variables sharing the same horizontal index.}.

The averaging kernels are constructed using the Gaspari--Cohn function, with centers evenly spaced along the vertical direction, and each with a half-width of $\num{10}$ levels. They are independently normalised in such a way that the natural variability of each observation matches that of the L96i variables. The $\num{8}$ non-normalised and normalised averaging kernels are displayed in \cref{fig:observation}. Finally, the observations are perturbed with a normal distribution of error covariance matrix $\mathbf{R}=\mathbf{I}$.

Note that this observation setup is sparse, as there are four time as many state variables as observations.

\subsubsection{The L2EnSRF-HML algorithm}

With non-local observations such as the ones described above, using DL only yields suboptimal results. Therefore, following the approach of \citet{farchi2019}, we include DL in the LEnSRF-HML, \cref{alg:alg-ensrf-hml}, but only in the horizontal direction. The resulting L$^2$EnSRF-HML algorithm uses DL in the horizontal direction (in which observation are local) and CL in the vertical direction (in which observations are non-local). The analysis is summarised in \cref{alg:alg-l2ensrf-hml}, in which several simplifications have been made.
\begin{itemize}
    \item In principle, four categories of parameters exist: global, horizontally local, vertically local, and both horizontally and vertically local. \Cref{alg:alg-l2ensrf-hml} only uses two categories: $\mathbf{q}$ gathers the set of horizontally local parameters and $\mathbf{p}$ the set of horizontally non-local parameters. In both categories, the parameters can be vertically local or not.
    \item Vertical localisation is performed using CL with the matrices $\boldsymbol{\rho}_{\mathsf{xx}}$, $\boldsymbol{\rho}_{\mathsf{px}}$ and $\boldsymbol{\rho}_{\mathsf{qx}}$.
    \item Horizontal localisation is performed using DL with $N_{\mathsf{h}}$ local analyses. Each local analysis updates the $h$-th column of state variables and model parameters, whose indices are written $c\left(h\right)$, using a single localisation matrix $\boldsymbol{\rho}_{h}$, common to state variables and model parameters.
\end{itemize}

For our estimation problem, we split the surrogate model coefficients into $\mathbf{p}$ and $\mathbf{q}$ as follows. The $\num{40}$ horizontal forcing coefficients $\mathbf{f}_\mathsf{h}$ are included in $\mathbf{q}$, and the $\num{17}$ monomial coefficients $\mathbf{a}$ are concatenated with the $\num{31}$ vertical forcing coefficients $\mathbf{f}_\mathsf{v}$ to form the horizontally non-local parameters $\mathbf{p}$:
\begin{equation}
    \mathbf{p} = \begin{bmatrix}\mathbf{a}\\
    \mathbf{f}_\mathsf{v}
    \end{bmatrix}.
\end{equation}

\begin{algorithm}
\caption{L$^2$EnSRF-HML analysis}
\label{alg:alg-l2ensrf-hml}
\begin{algorithmic}[1]
\renewcommand{\algorithmicrequire}{\textbf{Parameters:}}
\renewcommand{\algorithmicensure}{\textbf{Input:}}
\renewcommand{\algorithmiccomment}[1]{\hfill$\triangleright$~\textit{#1}}
\REQUIRE{Horizontal localisation matrices $\left\{\boldsymbol{\rho}_{h}, h=1, \ldots, N_{\mathsf{h}}\right\}$, vertical localisation matrices $\boldsymbol{\rho}_{\mathsf{xx}}$, $\boldsymbol{\rho}_{\mathsf{px}}$, and $\boldsymbol{\rho}_{\mathsf{qx}}$, tapering parameters $\zeta_{\mathsf{p}}$ and $\zeta_{\mathsf{q}}$}
\ENSURE{Forecast ensemble $\mathbf{E}^{\mathsf{f}}$}
\STATE{$\bar{\mathbf{z}}^{\mathsf{f}}=\mathbf{E}^{\mathsf{f}}\mathbf{1}/N_{\mathsf{e}}$}
\STATE{$\mathbf{Z}^{\mathsf{f}}=\left(\mathbf{E}^{\mathsf{f}}-\bar{\mathbf{z}}^{\mathsf{f}}\mathbf{1}^{\top}\right)/\sqrt{N_{\mathsf{e}}-1}$}
\STATE{$\widehat{\mathbf{Y}} = \boldsymbol{\mathcal{H}}\left(\mathbf{E}^{\mathsf{f}}\right)\left(\mathbf{I} - \mathbf{1}\mathbf{1}^\top/N_\mathsf{e}\right)/\sqrt{N_\mathsf{e}-1}$}
\STATE{$\mathbf{B}_{\mathsf{xx}}=\boldsymbol{\rho}_{\mathsf{xx}}\circ\left[\mathbf{Z}^{\mathsf{f}}_{\mathsf{x}}\left(\mathbf{Z}^{\mathsf{f}}_{\mathsf{x}}\right)^{\top}\right]$}
\STATE{$\mathbf{B}_{\mathsf{qx}}=\boldsymbol{\rho}_{\mathsf{qx}}\circ\left[\mathbf{Z}^{\mathsf{f}}_{\mathsf{q}}\left(\mathbf{Z}^{\mathsf{f}}_{\mathsf{x}}\right)^{\top}\right]$}
\STATE{$\mathbf{B}_{\mathsf{px}}=\boldsymbol{\rho}_{\mathsf{px}}\circ\left[\mathbf{Z}^{\mathsf{f}}_{\mathsf{p}}\left(\mathbf{Z}^{\mathsf{f}}_{\mathsf{x}}\right)^{\top}\right]$}
\FOR{$h=1$ \TO $N_{\mathsf{h}}$}
    \STATE{$\mathbf{R}^{-1}_{h}=\boldsymbol{\rho}_{h}\circ\mathbf{R}^{-1}$}
    \STATE{$\mathbf{T}_h = \mathbf{I+R}^{-1/2}_h\mathbf{H}_{\mathsf{x}}\mathbf{B}_{\mathsf{xx}}\mathbf{H}^{\top}_{\mathsf{x}}\mathbf{R}_h^{-1/2}$}
    \STATE{$\mathbf{u}_{\mathsf{x}} = \mathbf{H}^{\top}_{\mathsf{x}}\mathbf{R}_h^{-1/2}\mathbf{T}_h^{-1}\mathbf{R}_h^{-1/2}\left(\mathbf{y}-\boldsymbol{\mathcal{H}}_{\mathsf{x}}\left(\bar{\mathbf{x}}^{\mathsf{f}}\right)\right)$}
    \STATE{$\mathbf{U}_{\mathsf{x}} = -\mathbf{H}^{\top}_{\mathsf{x}}\mathbf{R}_h^{-1/2}\left(\mathbf{T}_h+\mathbf{T}_h^{1/2}\right)^{-1}\mathbf{R}_{h}^{-1/2}\widehat{\mathbf{Y}}$}
    \STATE{$\left[\Delta\bar{\mathbf{x}}\right]_{c(h)}=\left[\mathbf{B}_{\mathsf{xx}}\mathbf{u}_{\mathsf{x}}\right]_{c(h)}$}\COMMENT{state, mean update}
    \STATE{$\left[\Delta\bar{\mathbf{q}}\right]_{c(h)}=\zeta_{\mathsf{q}}\left[\mathbf{B}_{\mathsf{qx}}\mathbf{u}_{\mathsf{x}}\right]_{c(h)}$}\COMMENT{horizontally local parameters, mean update}

    \STATE{$\left[\Delta\mathbf{Z}_{\mathsf{x}}\right]_{{c(h)}, :}=\left[\mathbf{B}_{\mathsf{xx}}\mathbf{U}_{\mathsf{x}}\right]_{{c(h)}, :}$}\COMMENT{state, perturbation update}
    \STATE{$\left[\Delta\mathbf{Z}_{\mathsf{q}}\right]_{{c(h)}, :}=\zeta_{\mathsf{q}}\left[\mathbf{B}_{\mathsf{qx}}\mathbf{U}_{\mathsf{x}}\right]_{{c(h)}, :}$}\COMMENT{horizontally local parameters, perturbation update}
    
    \STATE{$\left[\mathbf{V}_{\mathsf{x}}\right]_{{c(h)}, :} = \left[\mathbf{U}_{\mathsf{x}}\right]_{{c(h)}, :}$}
    \STATE{$\left[\mathbf{v}_{\mathsf{x}}\right]_{c(h)} = \left[\mathbf{u}_{\mathsf{x}}\right]_{c(h)}$}
\ENDFOR
\STATE{$\Delta\bar{\mathbf{p}}=\zeta_{\mathsf{p}}\mathbf{B}_{\mathsf{px}}\mathbf{v}_{\mathsf{x}}$}\COMMENT{horizontally non-local parameters, mean update}
\STATE{$\Delta\mathbf{Z}_{\mathsf{p}}=\zeta_{\mathsf{p}}\mathbf{B}_{\mathsf{px}}\mathbf{V}_{\mathsf{x}}$}\COMMENT{horizontally non-local parameters, perturbation update}
\RETURN{$\mathbf{E}^{\mathsf{a}}=\left(\bar{\mathbf{z}}^{\mathsf{f}}+\Delta\bar{\mathbf{z}}\right)\mathbf{1}^{\top}+\sqrt{N_{\mathsf{e}}-1}\left(\mathbf{Z}^{\mathsf{f}}+\Delta\mathbf{Z}\right)$}\COMMENT{analysis ensemble}
\end{algorithmic}
\end{algorithm}

\subsubsection{Ensemble initialisation}

The ensemble initialisation is performed following the method described in \cref{sssec:illustration-1d-setup-ens-init}. The state is initialised with a standard deviation of $0.5$, smaller than in the one-dimensional test series because the time-averaged analysis error is expected to be smaller in the present experiment.
The standard deviations for the horizontal and vertical forcing coefficients have been set to $\num{0.17}$ and $\num{0.012}$, respectively, in such a way that the initial RMSE for the $\num{1280}$ reconstructed forcing coefficients (the outer product of the horizontal and vertical coefficients) ranges between $\num{0.15}$ and $\num{0.2}$.
Finally, the monomial coefficients initial standard deviation is set to $\sigma_\mathsf{a} = \num{0.1}$, once again smaller than in the one-dimensional test series as it makes the convergence faster.

\subsubsection{Algorithm parametrisation}

As seen in \cref{alg:alg-l2ensrf-hml}, the L$^2$EnSRF-HML analysis requires four kinds of localisation matrices.

First, the horizontal localisation matrices $\boldsymbol{\rho}_h$ are given by
\begin{equation}
    \left[\boldsymbol{\rho}_{h}\right]_{pq} \triangleq \sqrt{\gc\left(\frac{2d_\mathsf{h}\left(i, h\right)}{r_\mathsf{h}}\right)\gc\left(\frac{2d_\mathsf{h}\left(j, h\right)}{r_\mathsf{h}}\right)},
\end{equation}
where $d_\mathsf{h}\left(i, h\right)$ and $d_\mathsf{h}\left(j, h\right)$ are the horizontal (circular) distances between the $i$-th observation and the $h$-th column and between the $j$-th observation and the $h$-th column, respectively, and $r_\mathsf{h}$ is the horizontal localisation radius.

Second, the vertical localisation matrix between state variables, $\boldsymbol{\rho}_\mathsf{xx}$, is given by
\begin{equation}
    \label{eq:illustration-2d-def-rho-xx}
    \left[\boldsymbol{\rho}_{\mathsf{xx}}\right]_{mn} \triangleq \gc\left(\frac{2d_\mathsf{v}\left(m, n\right)}{r_\mathsf{v}}\right),
\end{equation}
where $d_\mathsf{v}\left(m, n\right)$ is the vertical distance between the $m$-th and $n$-th state variables and $r_\mathsf{v}$ is the vertical localisation radius.

Third, the vertical cross-localisation matrix between state variables and horizontally local parameters, $\boldsymbol{\rho}_\mathsf{qx}$ is set to $\boldsymbol{\Pi}$ because in this specific case the horizontally local parameters (the $\num{40}$ horizontal forcing coefficients) are not vertically local. 

Finally, the vertical cross-localisation matrix between state variables and horizontally non-local parameters, $\boldsymbol{\rho}_\mathsf{px}$ follows the same structure as the horizontally local parameters:
\begin{equation}
    \boldsymbol{\rho}_\mathsf{px} = \begin{bmatrix}\boldsymbol{\Pi} \\
 \boldsymbol{\rho}_{\mathsf{vx}}
    \end{bmatrix},
\end{equation}
where the first block, corresponding to the cross-localisation between the monomial coefficients $\mathbf{a}$ and the state variables, is set to $\boldsymbol{\Pi}$.
Because the monomial coefficients are global, this matrix should be row-wise uniform, but we additionally assume it to be fully uniform for the sake of simplicity. The second block $\boldsymbol{\rho}_{\mathsf{vx}}$, corresponding to the cross-localisation between the $N_{\mathsf{v}}-1=\num{31}$ vertical forcing coefficients and the state variables, is given by
\begin{equation}
    \left[\boldsymbol{\rho}_{\mathsf{vx}}\right]_{mn} \triangleq \gc\left(\frac{2d_\mathsf{v}\left(m, n\right)}{r_\mathsf{v}}\right).
\end{equation}
In this equation, $d_\mathsf{v}\left(m, n\right)$ is the vertical distance between the $m$-th vertical forcing coefficient and the $n$-th state variable, and $r_\mathsf{v}$ is the same vertical localisation radius as in \cref{eq:illustration-2d-def-rho-xx} to reduce the number of algorithmic parameters. Note that, by construction, the $m$-th vertical forcing coefficient has the same vertical location as the state variables within the $(m+1)$-th layer.

To summarise, the L$^2$EnSRF-HML analysis depends on two localisation radii $r_\mathsf{h}$ and $r_\mathsf{v}$. The analysis also depends on the two tapering coefficients $\zeta_{\mathsf{p}}$ and $\zeta_{\mathsf{q}}$. In order to further reduce the number of algorithmic parameters, and given the results of \cref{fig:LETKF-HML-2}, we set $\zeta_{\mathsf{p}}=\zeta_{\mathsf{q}}\triangleq\zeta$. Moreover, as in the one-dimensional test series, we use a multiplicative inflation on the prior with a uniform and constant in time coefficient $\lambda$. For each experiment, the algorithmic parameters $\left(r_\mathsf{h}, r_\mathsf{v}, \zeta, \lambda\right)$ are tuned so as to yield optimal scores.

\subsection{Results}

The goal of the present test series is to show that it is possible to estimate the parameters of the surrogate model alongside the state variables, but also that parameter localisation is efficient. To that purpose, we perform four types of experiments. We first test the EnSRF-ML and the L$^2$EnSRF-HML (with and without localisation). For comparison, we also test the EnSRF and the L$^2$EnSRF, with known model. The results are described in the following sections, and summarised in \cref{tab:result_loc}.

\begin{table}[tbp]
    %\rowcolors*{2}{}{}
    \footnotesize
    \centering
    \caption{Summary of the results for the two-dimensional test series with the mL96 model. For each experiment, we specify the inference problem (first column), the associated dimension of the unstable and neutral subspace of the augmented state dynamics $N_{0}$ (second column), the analysis algorithm (third column), the forecast model (fourth column), whether localisation is used or not (fifth column), the ensemble size $N_{\mathsf{e}}$ (seventh column), and the time-averaged state analysis RMSE (last column). The symbol $\geq$ in the ensemble size column means that increasing the ensemble size does not yield significantly better scores.}
    \label{tab:result_loc}
\begin{tabular}{clllclr}
\hline
Inference problem
& $N_{0}$
& Algorithm
& Model
& Loc.
& $N_{\mathsf{e}}$
& state RMSE \\ \hline
\multirow{2}{*}{1: $\mathbf{x}$}
& \multirow{2}{*}{$\approx\num{50}$}
& EnSRF
& mL96
&
& $\geq\num{50}$
& $\num{0.08}$ \\
& & L$^{2}$EnSRF
& mL96
& \checkmark
& $\geq\num{10}$
& $\num{0.08}$ \\ \hline
\multirow{2}{*}{2: $\mathbf{\left(\mathbf{x}, \mathbf{a}, \mathbf{f}_{\mathsf{v}}, \mathbf{f}_{\mathsf{h}}\right)}$}
& \multirow{2}{*}{$\approx50+88$}
& EnSRF-HML
& $\sur\left(\mathbf{a}, \mathbf{f}_{\mathsf{v}}, \mathbf{f}_{\mathsf{h}}\right)$
&
& $\geq\num{140}$
& $\num{0.11}$ \\
& & L$^{2}$EnSRF-HML
& $\sur\left(\mathbf{a}, \mathbf{f}_{\mathsf{v}}, \mathbf{f}_{\mathsf{h}}\right)$
& \checkmark
& $\num{50}$
& $\num{0.12}$ \\ \hline
\end{tabular}
\end{table}

\subsubsection{Estimation of the state variables only}

With known model, and without localisation, $\num{50}$ ensemble members are necessary to accurately estimate the $\num{1280}$ state variables only, which corresponds more or less to the dimension of the unstable and neutral subspace \citep{bocquet2017}. The time-averaged analysis RMSE is around $\num{0.08}$. With localisation, only $\num{8}$ ensemble members \citep{farchi2021} are required for a successful estimation, and the best scores require an ensemble of $\num{10}$ members.

\subsubsection{Estimation of both state variables and model parameters without localisation}
\label{sssec:illustration-2d-results-hml-without-loc}

Without localisation, an ensemble of $\num{140}$ members is sufficient to accurately estimate the $\num{88}$ model parameters alongside the $\num{1280}$ state variables. The time-averaged state analysis is around $\num{0.11}$. Although this experiment is not the main result of the present two-dimensional test series, we think that it provides a reasonable approximation of the best scores that can be obtained with the L$^2$EnSRF-HML. 

More precisely, we have found that the state analysis RMSE decreases almost linearly as the ensemble size $N_{\mathsf{e}}$ increases (not shown here). The score stops improving when $N_{\mathsf{e}}$ reaches a critical value, around $N_{\mathsf{e}}=\num{140}$ here, which is close to the dimension of the unstable and neutral subspace of the augmented state dynamics \citep{bocquet2021}. When the ensemble size $N_{\mathsf{e}}$ is close, but smaller to $\num{140}$, the algorithm is able to accurately estimate the monomial coefficients $\mathbf{a}$ but struggles to estimate the forcing coefficients $\mathbf{f}_{\mathsf{v}}$ and $\mathbf{f}_{\mathsf{h}}$. This is once again related to the fact that the surrogate model is more sensitive to perturbations of $\mathbf{a}$.

\subsubsection{Estimation of both state variables and model parameters with localisation}

With localisation, an ensemble of only $\num{50}$ members is sufficient to accurately estimate the $\num{88}$ model parameters alongside the state. From the previous experiments, we know that $\num{10}$ members are sufficient to estimate the state variables. Additionally, $\num{17}$ members are required to estimate the $\num{17}$ monomial coefficients $\mathbf{a}$, which are neutral modes of the augmented state dynamics and which do not benefit from localisation. This means that only about $50-10-17=\num{23}$ additional members are necessary to estimate the $40+31=\num{71}$ horizontal and vertical forcing coefficients $\mathbf{f}_{\mathsf{v}}$ and $\mathbf{f}_{\mathsf{h}}$. This shows that parameter localisation is indeed effective.

\begin{figure}[htbp]
\centering
\includegraphics{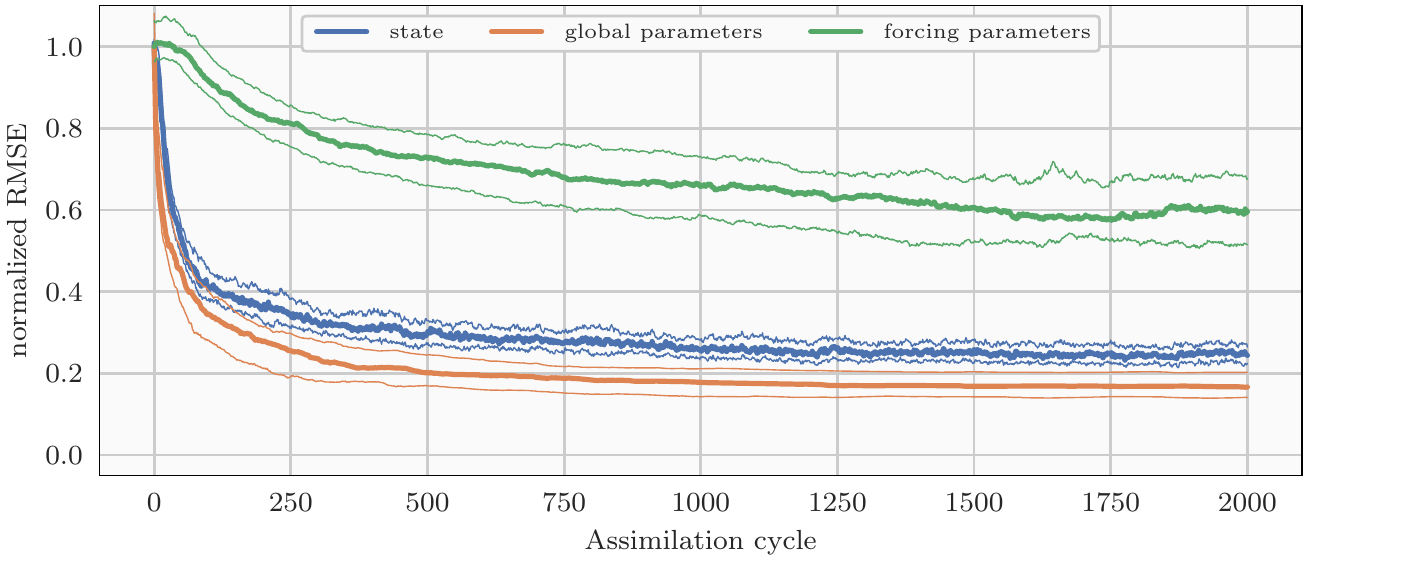}

\caption{Time series of instantaneous analysis RMSE for the two-dimensional test series (with the mL96 model) using the L$^2$EnSRF-HML. The state RMSE is shown in blue, the RMSE for the $\num{17}$ monomial coefficients $\mathbf{a}$ in yellow, and the RMSE for the $\num{1280}$ reconstructed forcing coefficients (the outer product of the horizontal and vertical coefficients) in green. In all cases, the RMSE is normalised by its initial value. The thick lines represent the median over $\num{100}$ experiments, and the thin lines represent the $32$-th and $68$-th percentiles.}
\label{fig:time-LLEnSRF-HML_loc}
\end{figure}

For this inference problem, \cref{fig:time-LLEnSRF-HML_loc} shows the results of an experiment with $\num{50}$ members. The conclusions are overall very similar to those in \cref{sec:all_coeffs_1D}. First, the experiment can be qualified as successful: after a spin-up period of several hundreds of cycles, the state analysis RMSE stabilises around $\num{0.12}$. Second, the improvement of the analysis is rather slow, once again because the algorithmic parameters have been chosen to minimise the asymptotic analysis error. Third, the different components of the augmented state are learnt on different time scales: the algorithm first corrects the state and the monomial coefficients $\mathbf{a}$, which are the most sensitive parameters. Finally, note that the time-averaged state analysis RMSE is a bit higher here ($\num{0.12}$) than in \cref{sssec:illustration-2d-results-hml-without-loc} without localisation ($\num{0.11}$), but we have checked that better scores can be obtained with localisation when using larger ensembles.

%--------------------------------------------------

\section{Conclusions}
\label{sec:conclusions}

In the wake of \citet{bocquet2021}, we have shown how the classical LETKF and LEnSRF can be generalised to estimate model parameters, both global and local, alongside the state variables. The assimilation of local parameters is natural with DL (\textit{i.e.} the LETKF), especially when model parameters and state variables are co-located. By contrast, CL (\textit{i.e.} the LEnSRF) is more suited than DL to the estimation of global parameters, and to the assimilation of non-local observations, at the cost of having to perform linear algebra in the whole state space. Introducing the ancillary variables defined in \cref{eq:alg-ensrf-ml-def-u-mean} and \cref{eq:alg-ensrf-ml-def-u-pert} for the LEnSRF-HML, and defined in \cref{eq:alg-etkf-ml-def-u-mean} and \cref{eq:alg-etkf-ml-def-u-pert} for the LETKF-HML, we have optimised these algorithms in such a way that it is not necessary to compute $\mathbf{B}_\mathsf{xx}^{-1}$ when evaluating the global parameter update from the local state update. Moreover, we have shown how to rigorously assimilate global parameters within the DL-based LETKF, assuming the observations are local. The existing and proposed algorithms are summarised in \cref{tab:enkf-ml-family}.

Introducing the L96i, an inhomogeneous variant of the L96 model, we have numerically tested the LETKF-HML and the LEnSRF-HML. The results are overall consistent with those of \citet{bocquet2021}, they show that the algorithms are able to learn the dynamics of a fully parametrised surrogate model alongside the state variables, and that parameter localisation is beneficial, in the sense that it is possible to estimate a large amount of parameters with reasonable ensemble sizes. 

Finally, we have generalised the L$^2$EnSRF algorithm proposed in \citet{farchi2021} to estimate model parameters alongside the state variables. The resulting algorithm, called L$^2$EnSRF-HML, combines DL in the horizontal direction and CL in the vertical direction, and is able to take into account the local nature of state variables and model parameters in both the horizontal and vertical directions. We have illustrated with success the algorithm using a challenging multilayer L96 experiment with radiance-like, non-local observations, in which the surrogate model has global monomial coefficients as well as vertically local and horizontally local forcing coefficients.

Several researchs could be initiated following this paper. First, we showed that using non-adaptive algorithmic parameters can be critical because the optimal algorithmic parameters change over time, as the surrogate model progressively improves; hence, adaptive algorithms should be investigated. Second, we have not demonstrated a clear practical advantage of the LETKF-ML over the empirical LETKF-Aksoy. We suspect that applying the methods to well designed sparsely observed systems could unveil accuracy differences.  Third, the online characteristics of the proposed algorithms could be exploited in situations with slow parameter time evolution, while offline algorithms would be unfit to this task. Finally, the next goal would be to learn the dynamics of more complex and non identifiable surrogate representations, with a more realistic dynamical model (both in term of physical interpretation and of state dimension).

\appendix
\section{General state variables and global parameter covariance matix} 
\label{app:general_zeta}
In this appendix, we focus on the global parameter problem and consider the most general cross localisation matrix given by \cref{eq:general-cross-localisation}. We would like to improve on Section 3.2.4 of \citet{bocquet2021} where the global parameters have been considered statistically homogeneous, such that $\boldsymbol{\zeta}_\mathsf{p} = \zeta_\mathsf{p} \mathbf{1}_\mathsf{p}$ was posed. The most general form of the localisation matrix is then:
\begin{equation}
       \label{eq:split-rho}
    \boldsymbol{\rho} = \begin{bmatrix}
    \,\boldsymbol{\rho}_{\mathsf{xx}} & \mathbf{1}_{\mathsf{x}} \boldsymbol{\zeta}^\top_\mathsf{p} \, \\ 
    \, \boldsymbol{\zeta}_\mathsf{p} \mathbf{1}^\top_{\mathsf{x}} & \boldsymbol{\rho}_{\mathsf{pp}} \, \\
    \end{bmatrix}.
\end{equation}
For $\boldsymbol{\rho}$ to be a correlation matrix, this implies constraints on the vector of tapering coefficients $\boldsymbol{\zeta}_\mathsf{p}$.
Assuming $\boldsymbol{\rho}_{\mathsf{xx}}$ is positive definite, $\boldsymbol{\rho}$ is positive semi-definite if and only if the Schur complement
\begin{equation}
\mathbf{S}_{\mathsf{pp}} = \boldsymbol{\rho}_{\mathsf{pp}} - \boldsymbol{\zeta}_\mathsf{p} \mathbf{1}^\top_{\mathsf{x}} \boldsymbol{\rho}_{\mathsf{xx}}^{-1} \mathbf{1}_\mathsf{x} \boldsymbol{\zeta}^\top_\mathsf{p}
\end{equation}
is positive semi-definite.

Alternatively, assuming $\boldsymbol{\rho}_{\mathsf{pp}}$ is positive definite, $\boldsymbol{\rho}$ is positive semi-definite if and only if the Schur complement
\begin{equation}
\mathbf{S}_{\mathsf{xx}} = \boldsymbol{\rho}_{\mathsf{xx}} - \mathbf{1}_\mathsf{x} \boldsymbol{\zeta}^\top_\mathsf{p}  \boldsymbol{\rho}_{\mathsf{pp}}^{-1} \boldsymbol{\zeta}_\mathsf{p} \mathbf{1}^\top_{\mathsf{x}}
\end{equation}
is positive semi-definite. Both conditions can be re-arranged as
\begin{equation}
\label{eq:schur_complement_pp}
\mathbf{S}_{\mathsf{pp}} = \boldsymbol{\rho}_{\mathsf{pp}} - \mathbf{1}_{\mathsf{x}}^\top\boldsymbol{\rho}_{\mathsf{xx}}^{-1} \mathbf{1}_{\mathsf{x}} \cdot
\boldsymbol{\zeta}_{\mathsf{p}}\boldsymbol{\zeta}_{\mathsf{p}}^\top ,
\end{equation}
and 
\begin{equation}
\label{eq:schur_complement_xx}
\mathbf{S}_{\mathsf{xx}} = \boldsymbol{\rho}_{\mathsf{xx}} -
\boldsymbol{\zeta}_{\mathsf{p}}^\top \boldsymbol{\rho}_{\mathsf{pp}}^{-1} \boldsymbol{\zeta}_{\mathsf{p}} \cdot
\boldsymbol{\Pi}_{\mathsf{xx}} .
\end{equation}
Since the second term on the right-hand-side of \cref{eq:schur_complement_pp} is of rank one, we only need to ensure that the condition is true along $\boldsymbol{\zeta}_{\mathsf{pp}}$:
\begin{equation}
     \boldsymbol{\zeta}_{\mathsf{p}}^\top \mathbf{S}_{\mathsf{pp}}\boldsymbol{\zeta}_{\mathsf{p}} = \boldsymbol{\zeta}_{\mathsf{p}}^\top \boldsymbol{\rho}_{\mathsf{pp}}\boldsymbol{\zeta}_{\mathsf{p}} - \mathbf{1}_{\mathsf{x}}^\top\boldsymbol{\rho}_{\mathsf{xx}}^{-1} \mathbf{1}_{\mathsf{x}} \cdot
\| \boldsymbol{\zeta}_{\mathsf{p}}\|^4 \ge 0 ,
\end{equation}
where $\| \cdot \|$ is the Euclidean norm.    
    
Likewise, since the second term on the right-hand-side of \cref{eq:schur_complement_xx} is of rank one, we only need to ensure that the condition is true along $\mathbf{1}_\mathsf{x}$:
\begin{equation}
\mathbf{1}_{\mathsf{x}}^\top \mathbf{S}_{\mathsf{xx}} \mathbf{1}_\mathsf{x} = \mathbf{1}_{\mathsf{x}}^\top  \boldsymbol{\rho}_{\mathsf{xx}} \mathbf{1}_\mathsf{x}  -
\boldsymbol{\zeta}_{\mathsf{p}}^\top \boldsymbol{\rho}_{\mathsf{pp}}^{-1} \boldsymbol{\zeta}_{\mathsf{p}} N^2_{\mathsf{x}} \ge 0 .
\end{equation}

As a consequence, a sufficient condition for the positivity of $\boldsymbol{\rho}$  that can be easily derived from \emph{both} of these scalar conditions, is 
\begin{equation}
    \| \boldsymbol{\zeta}_{\mathsf{p}}\| \le \sqrt{\frac{\lambda_\mathsf{p}^\mathrm{min}\lambda_\mathsf{x}^\mathrm{min}}{N_{\mathsf{x}}}} ,
\end{equation}
where $\lambda_\mathsf{p}^\mathrm{min}, \lambda_\mathsf{x}^\mathrm{min} $ are the smallest eigenvalues of $\boldsymbol{\rho}_{\mathsf{pp}}, \boldsymbol{\rho}_{\mathsf{xx}}$, respectively. 
This result includes the more heuristic and less general equation (26) of \citet{bocquet2021}.
This upper bound is very likely to be suboptimal but it suggests that it scales in the state space dimension 
as $N_{\mathsf{x}}^{-1/2}$.

\section{Proof of the corrected ETKF-ML formulae} 
\label{app:proof-corrected-etkf-ml-formulae}

In this appendix, we provide a proof of \cref{eq:alg-etkf-ml-param-update,eq:alg-etkf-ml-def-u}, which is the core of the ETKF-ML parameter update. Let us start with $\Delta\bar{\mathbf{p}}$. From \cref{eq:alg-ensrf-ml-def-u-mean,eq:alg-ensrf-ml-param-mean-update}, we have
\begin{align}
    \Delta\bar{\mathbf{p}} &= \mathbf{B}_{\mathsf{px}}\mathbf{u}_{\mathsf{x}},\\
    &= \mathbf{Z}^{\mathsf{f}}_{\mathsf{p}}\left(\mathbf{Z}^{\mathsf{f}}_{\mathsf{x}}\right)^{\top}\mathbf{u}_{\mathsf{x}},\\
    &= \mathbf{Z}^{\mathsf{f}}_{\mathsf{p}}\left(\mathbf{Z}^{\mathsf{f}}_{\mathsf{x}}\right)^{\top}\mathbf{H}^{\top}_{\mathsf{x}}\mathbf{R}^{-1/2}\mathbf{T}_{\mathsf{y}}^{-1}\mathbf{R}^{-1/2}\left(\mathbf{y}-\boldsymbol{\mathcal{H}}\left(\bar{\mathbf{z}}^{\mathsf{f}}\right)\right),\\
\end{align}
where $\mathbf{T}_{\mathsf{y}}$ is given by \cref{eq:alg-ensrf-ml-Ty}:
\begin{equation}
    \mathbf{T}_{\mathsf{y}} = \mathbf{I}+\mathbf{R}^{-1/2}\mathbf{H}_{\mathsf{x}}\mathbf{Z}^{\mathsf{f}}_{\mathsf{x}}\left(\mathbf{Z}^{\mathsf{f}}_{\mathsf{x}}\right)^{\top}\mathbf{H}^{\top}_{\mathsf{x}}\mathbf{R}^{-1/2}.
\end{equation}
Using the approximation $\mathbf{H}_{\mathsf{x}}\mathbf{Z}^{\mathsf{f}}_{\mathsf{x}}=\mathbf{H}\mathbf{Z}^{\mathsf{f}}\approx\boldsymbol{\mathcal{H}}\left(\mathbf{Z}^{\mathsf{f}}\right)=\mathbf{R}^{1/2}\mathbf{Y}$, we get
\begin{align}
    \Delta\bar{\mathbf{p}} &= \mathbf{Z}^{\mathsf{f}}_{\mathsf{p}}\mathbf{Y}^{\top}\mathbf{T}_{\mathsf{y}}^{-1}\mathbf{R}^{-1/2}\left(\mathbf{y}-\boldsymbol{\mathcal{H}}\left(\bar{\mathbf{z}}^{\mathsf{f}}\right)\right),\\
    \mathbf{T}_{\mathsf{y}} &= \mathbf{I}+\mathbf{Y}\mathbf{Y}^{\top}.
\end{align}
In addition, we have
\begin{align}
    \mathbf{T}^{-1}_{\mathsf{y}} &= \left(\mathbf{I}+\mathbf{Y}\mathbf{Y}^{\top}\right)^{-1},\\
    &= \left(\mathbf{I}+\mathbf{Y}\mathbf{Y}^{\top}\right)^{-1}\left(\mathbf{I}+\mathbf{Y}\mathbf{Y}^{\top}-\mathbf{Y}\mathbf{Y}^{\top}\right),\\
    &= \mathbf{I} - \left(\mathbf{I}+\mathbf{Y}\mathbf{Y}^{\top}\right)^{-1}\mathbf{Y}\mathbf{Y}^{\top},\label{eq:appendix-proof-u-mean-before-sml}\\
    &= \mathbf{I} - \mathbf{Y}\left(\mathbf{I}+\mathbf{Y}^{\top}\mathbf{Y}\right)^{-1}\mathbf{Y}^{\top},\label{eq:appendix-proof-u-mean-after-sml}\\
    &= \mathbf{I} - \mathbf{Y}\mathbf{T}^{-1}_{\mathsf{e}}\mathbf{Y}^{\top},
\end{align}
where \cref{eq:appendix-proof-u-mean-after-sml} is obtained from \cref{eq:appendix-proof-u-mean-before-sml} using the matrix shift lemma. We conclude that
\begin{align}
    \Delta\bar{\mathbf{p}} &= \mathbf{Z}^{\mathsf{f}}_{\mathsf{p}}\mathbf{Y}^{\top}\left(\mathbf{I}-\mathbf{Y}\mathbf{T}^{-1}_{\mathsf{e}}\mathbf{Y}^{\top}\right)\mathbf{R}^{-1/2}\left(\mathbf{y}-\boldsymbol{\mathcal{H}}\left(\bar{\mathbf{z}}^{\mathsf{f}}\right)\right),\\
    &= \mathbf{Z}^{\mathsf{f}}_{\mathsf{p}}\mathbf{Y}^{\top}\left\{
    \mathbf{R}^{-1/2}\left(\mathbf{y}-\boldsymbol{\mathcal{H}}\left(\bar{\mathbf{z}}^{\mathsf{f}}\right)\right)-\mathbf{Y}\mathbf{T}^{-1}_{\mathsf{e}}\mathbf{Y}^{\top}\mathbf{R}^{-1/2}\left(\mathbf{y}-\boldsymbol{\mathcal{H}}\left(\bar{\mathbf{z}}^{\mathsf{f}}\right)\right)\right\},\\
    &= \mathbf{Z}^{\mathsf{f}}_{\mathsf{p}}\mathbf{Y}^{\top}\left\{
    \mathbf{R}^{-1/2}\left(\mathbf{y}-\boldsymbol{\mathcal{H}}\left(\bar{\mathbf{z}}^{\mathsf{f}}\right)\right)-\mathbf{Y}\mathbf{w}^{\mathsf{a}}\right\},
\end{align}
which implies \cref{eq:alg-etkf-ml-param-update-mean,eq:alg-etkf-ml-def-u-mean}.

Likewise, from \cref{eq:alg-ensrf-ml-def-u-pert,eq:alg-ensrf-ml-param-pert-update}, we have
\begin{align}
    \Delta\mathbf{Z}_{\mathsf{p}} &= \mathbf{B}_{\mathsf{px}}\mathbf{U}_{\mathsf{x}},\\
    &= -\mathbf{Z}^{\mathsf{f}}_{\mathsf{p}}\left(\mathbf{Z}^{\mathsf{f}}_{\mathsf{x}}\right)^{\top}\mathbf{H}_{\mathsf{x}}^{\top}\mathbf{R}^{-1/2}\left(\mathbf{T}_{\mathsf{y}}+\mathbf{T}_{\mathsf{y}}^{1/2}\right)^{-1}\mathbf{R}^{-1/2}\boldsymbol{\mathcal{H}}\left(\mathbf{Z}^{\mathsf{f}}\right),\\
    &= -\mathbf{Z}^{\mathsf{f}}_{\mathsf{p}}\mathbf{Y}^{\top}\left(\mathbf{I}+\mathbf{YY}^{\top}+\left(\mathbf{I}+\mathbf{YY}^{\top}\right)^{1/2}\right)^{-1}\mathbf{Y},\label{eq:appendix-proof-u-pert-before-sml}\\
    &= -\mathbf{Z}^{\mathsf{f}}_{\mathsf{p}}\mathbf{Y}^{\top}\mathbf{Y}\left(\mathbf{I}+\mathbf{Y}^{\top}\mathbf{Y}+\left(\mathbf{I}+\mathbf{Y}^{\top}\mathbf{Y}\right)^{1/2}\right)^{-1},\label{eq:appendix-proof-u-pert-after-sml}\\
    &= -\mathbf{Z}^{\mathsf{f}}_{\mathsf{p}}\mathbf{Y}^{\top}\mathbf{Y}\left(\mathbf{T}_{\mathsf{e}}+\mathbf{T}_{\mathsf{e}}^{1/2}\right)^{-1},
\end{align}
which implies \cref{eq:alg-etkf-ml-param-update-pert,eq:alg-etkf-ml-def-u-pert}. Note that \cref{eq:appendix-proof-u-pert-after-sml} is obtained from \cref{eq:appendix-proof-u-pert-before-sml} using once again the matrix shift lemma.

\section{A variant of the LEnSRF-HML without the adjoint of the observation operator}
\label{sec:variant-EnSRF-HML}

In this section, we assume the observation operator to be fully local, as defined in \cref{sssec:alg-etkf-hml-analysis}. Following  \citet{asch2016}, section 6.5.1.3, we can use the map $h$ to transpose the localisation matrix $\boldsymbol{\rho}_{\mathsf{xx}}$ from $N_\mathsf{x} \times N_\mathsf{x}$ to
$N_\mathsf{x} \times N_\mathsf{y}$ and $N_\mathsf{y} \times N_\mathsf{y}$ via, e.g., $[\boldsymbol{\rho}_{\mathsf{xy}}]_{n,p} = [\boldsymbol{\rho}_{\mathsf{xx}}]_{n,h(p)}$, or $[\boldsymbol{\rho}_{\mathsf{yy}}]_{p,q} = [\boldsymbol{\rho}_{\mathsf{xx}}]_{h(p),h(q)}$.
This unequivocally defines $\boldsymbol{\rho}_{\mathsf{yy}}$ and $\boldsymbol{\rho}_{\mathsf{xy}}$. By construction, the tangent linear $\mathbf{H}_{\mathsf{x}}\in\mathbb{R}^{N_{\mathsf{y}}\times N_{\mathsf{x}}}$ of $\boldsymbol{\mathcal{H}}_{\mathsf{x}}$ satisfies $\left[\mathbf{H} \right]_{p,n}=0$ if $n \neq h(p)$, in such a way that for any matrix $\mathbf{M}\in\mathbb{R}^{N_{\mathsf{x}}\times N_{\mathsf{x}}}$, we have:
\begin{align}
	\left(\boldsymbol{\rho}_{\mathsf{xx}} \circ \mathbf{M}\right)\mathbf{H}^\top &= \boldsymbol{\rho}_{\mathsf{xy}} \circ \left(\mathbf{M}\mathbf{H}^\top \right),\\
		\mathbf{H}\left(\boldsymbol{\rho}_{\mathsf{xx}} \circ \mathbf{M}\right)\mathbf{H}^\top &= \boldsymbol{\rho}_{\mathsf{yy}} \circ \left(\mathbf{H}\mathbf{M}\mathbf{H}^\top \right).
\end{align}
In addition, we make the assumption $\mathbf{H}_{\mathsf{x}}\mathbf{Z}^{\mathsf{f}}_{\mathsf{x}}=\mathbf{H}\mathbf{Z}^{\mathsf{f}}\approx\boldsymbol{\mathcal{H}}\left(\mathbf{Z}^{\mathsf{f}}\right)$, as in \cref{app:proof-corrected-etkf-ml-formulae}.

With these two elements, we can rewrite \cref{alg:alg-ensrf-hml} into \cref{alg:alg-ensrf-hml-yy},  \textit{i.e.} with ancillary increments  expressed in the observation space and covariance localisation performed in the observation space. 

\begin{algorithm}
\caption{LEnSRF-HML analysis for a fully local observation operator}
\label{alg:alg-ensrf-hml-yy}
\begin{algorithmic}[1]
\renewcommand{\algorithmicrequire}{\textbf{Parameters:}}
\renewcommand{\algorithmicensure}{\textbf{Input:}}
\renewcommand{\algorithmiccomment}[1]{\hfill$\triangleright$~\textit{#1}}
\REQUIRE{localisation matrices $\boldsymbol{\rho}_{\mathsf{xx}}$ and $\boldsymbol{\rho}_{\mathsf{qx}}$, tapering parameters $\zeta_{\mathsf{p}}$ and $\zeta_{\mathsf{q}}$; the matrices $\boldsymbol{\rho}_{\mathsf{yy}}$, $\boldsymbol{\rho}_{\mathsf{xy}}$ and $\boldsymbol{\rho}_{\mathsf{qy}}$ are inferred from $\boldsymbol{\rho}_{\mathsf{xx}}$ and $\boldsymbol{\rho}_{\mathsf{qx}}$ using the locality of the observation operator.}
\ENSURE{Forecast ensemble $\mathbf{E}^{\mathsf{f}}$}
\STATE{$\bar{\mathbf{z}}^{\mathsf{f}}=\mathbf{E}^{\mathsf{f}}\mathbf{1}/N_{\mathsf{e}}$}
\STATE{$\mathbf{Z}^{\mathsf{f}}=\left(\mathbf{E}^{\mathsf{f}}-\bar{\mathbf{z}}^{\mathsf{f}}\mathbf{1}^{\top}\right)/\sqrt{N_{\mathsf{e}}-1}$}
\STATE{$\mathbf{Y}= \mathbf{R}^{-1/2}\boldsymbol{\mathcal{H}}\left(\mathbf{E}^{\mathsf{f}}\right)\left(\mathbf{I} - \mathbf{1}\mathbf{1}^\top/N_\mathsf{e}\right)/\sqrt{N_\mathsf{e}-1}$}
\STATE{$\mathbf{B}_{\mathsf{yy}}=\boldsymbol{\rho}_{\mathsf{yy}}\circ\left(\mathbf{Y}\mathbf{Y}^{\top}\right)$}
\STATE{$\mathbf{B}_{\mathsf{xy}}=\boldsymbol{\rho}_{\mathsf{xy}}\circ\left(\mathbf{Z}^{\mathsf{f}}_{\mathsf{x}}\mathbf{Y}^{\top}\right)$}
\STATE{$\mathbf{B}_{\mathsf{qy}}=\boldsymbol{\rho}_{\mathsf{qy}}\circ\left(\mathbf{Z}^{\mathsf{f}}_{\mathsf{q}}\mathbf{Y}^{\top}\right)$}
\STATE{$\mathbf{B}_{\mathsf{py}}=\mathbf{Z}^{\mathsf{f}}_{\mathsf{p}}\mathbf{Y}^{\top}$}
\STATE{$\mathbf{T}_\mathsf{y} = \mathbf{I} + \mathbf{B}_{\mathsf{yy}}$}
\STATE{$\mathbf{u}_{\mathsf{y}} = \mathbf{T}_\mathsf{y}^{-1}\mathbf{R}^{-1/2}\left(\mathbf{y}-\boldsymbol{\mathcal{H}}_{\mathsf{x}}\left(\bar{\mathbf{x}}^{\mathsf{f}}\right)\right)$}
\STATE{$\mathbf{U}_{\mathsf{y}} = -\left(\mathbf{T}_\mathsf{y}+\mathbf{T}_\mathsf{y}^{1/2}\right)^{-1}\mathbf{Y}$}
\STATE{$\Delta\bar{\mathbf{x}}=\mathbf{B}_{\mathsf{xy}}\mathbf{u}_{\mathsf{y}}$}\COMMENT{state, mean update}
\STATE{$\Delta\bar{\mathbf{q}}=\zeta_{\mathsf{q}}\mathbf{B}_{\mathsf{qy}}\mathbf{u}_{\mathsf{y}}$}\COMMENT{local parameters, mean update}
\STATE{$\Delta\bar{\mathbf{p}}=\zeta_{\mathsf{p}}\mathbf{B}_{\mathsf{py}}\mathbf{u}_{\mathsf{y}}$}\COMMENT{global parameters, mean update}
\STATE{$\Delta\mathbf{Z}_{\mathsf{x}}=\mathbf{B}_{\mathsf{xy}}\mathbf{U}_{\mathsf{y}}$}\COMMENT{state, perturbation update}
\STATE{$\Delta\mathbf{Z}_{\mathsf{q}}=\zeta_{\mathsf{q}}\mathbf{B}_{\mathsf{qy}}\mathbf{U}_{\mathsf{y}}$}\COMMENT{local parameters, perturbation update}
\STATE{$\Delta\mathbf{Z}_{\mathsf{p}}=\zeta_{\mathsf{p}}\mathbf{B}_{\mathsf{py}}\mathbf{U}_{\mathsf{y}}$}\COMMENT{global parameters, perturbation update}
\RETURN{$\mathbf{E}^{\mathsf{a}}=\left(\bar{\mathbf{z}}^{\mathsf{f}}+\Delta\bar{\mathbf{z}}\right)\mathbf{1}^{\top}+\sqrt{N_{\mathsf{e}}-1}\left(\mathbf{Z}^{\mathsf{f}}+\Delta\mathbf{Z}\right)$}\COMMENT{analysis ensemble}
\end{algorithmic}
\end{algorithm}

\section{Equations of the surrogate model}
\label{sec:surrogate-model}

Similarly to the L96 model, the surrogate model is defined by a set of ODEs over a periodic domain with $N_{\mathsf{x}}$ variables, indexed by $n = 1, \dots,  N_{\mathsf{x}}$:
\begin{align}
       \frac{\mathrm{d}x_n}{\mathrm{d}t} = \underbrace{\vphantom{\sum_{l=0}^{L}\sum_{m=-L}^{L-l} M_{l+1, L+m+1}x_{n+m} x_{n+m+l}}\sum_{m=-L}^L v_{L+m+1} x_{n+m}}_{\text{linear}} + \underbrace{\sum_{l=0}^{L}\sum_{m=-L}^{L-l} M_{l+1, L+m+1}x_{n+m} x_{n+m+l}}_{\text{quadratic}} + \underbrace{\vphantom{\sum_{l=0}^{L}\sum_{m=-L}^{L-l} M_{l+1, L+m+1}x_{n+m} x_{n+m+l}}f_n}_{\text{forcing}},
\end{align}
where: the vector $\mathbf{v}$ contains the linear coefficients, the relevant entries of $\mathbf{M}$ are the quadratic coefficients, and $\mathbf{f}$ contains the local forcing coefficients. The entries of $\mathbf{v}$ and the relevant entries of $\mathbf{M}$ are global parameters and concatenated in a vector $\mathbf{a}$, while the local parameters are contained in $\mathbf{f}$.

To be more specific, we consider the parameters $\mathbf{v}$ and $\mathbf{M}$ as global, because every state variable $x_n$ tendency depends on each of those parameters, in the same way and independently from the state variable location. We consider the parameters $\mathbf{f}$ as local because the tendencies of every state variable does not depend in the same way with each of those parameters, and more precisely, in this case, each forcing parameter $f_n$ only affect the tendencies of one state variable $x_n$.

The tendencies of a given state variable $x_n$ does only depends on the state variables $x_m, \lvert n-m\rvert \leq L$, where $L$ is the stencil radius. The dynamic of the model is therefore \emph{local}. The bilinear term affects the tendencies of a state variable $x_n$ through product of couples of state variables $x_m x_p, \lvert n-m\rvert \leq L, \lvert n-p\rvert \leq L, \lvert m-p\rvert \leq L$.

Higher order and range of dependencies than those of the tendencies are generated through time integration.

\section{Summary of the presented algorithms} 
\label{sec:resume-algo}

The local EnKF-based methods for estimating state, global and both local and global parameters are summarised in \cref{tab:enkf-ml-family}.

\begin{table}[tbp]
    %\rowcolors*{2}{}{}
    %\footnotesize
    \scriptsize
    \centering
    \caption{Summary of the EnKF-ML family of algorithms}
    \label{tab:enkf-ml-family}
\begin{tabular}{cccc}
\hline
Inference problem
& Dom. Local.
& Cov. Local.
& Dom. + Cov. Local.
\\
& \textcolor{snsred}{local obs. only}
& \textcolor{snsred}{numerically costly}
&
\\ \hline
State
& LETKF \citep{hunt2007}
& LEnSRF \citep{whitaker2002}
& L$^{2}$EnSRF \citep{farchi2019}
\\ \hline
State
& LETKF-ML \citep{bocquet2021}
& LEnSRF-ML \citep{bocquet2021}
& L$^{2}$EnSRF-ML
\\
+ global param.
& \textcolor{snsblue}{new implementation}
& \textcolor{snsblue}{new implementation}
& \textcolor{snsred}{not discussed}
\\ \hline
State
& LETKF-HML
& LEnSRF-HML
& L$^{2}$EnSRF-HML
\\
+ global \& local param.
& \textcolor{snsgreen}{new algorithm}
& \textcolor{snsgreen}{new algorithm}
& \textcolor{snsgreen}{new algorithm}
\\ \hline

\end{tabular}
\end{table}

\section*{acknowledgements}
CEREA is a member of Institut Pierre-Simon Laplace (IPSL). The authors are thankful to two anonymous reviewers for their time and very insightful remarks.

\section*{conflict of interest}
None.

\section*{Supporting Information}
None.

%\printendnotes

\bibliography{paper}
\end{document}